\documentclass[lettersize,journal]{IEEEtran}
\usepackage{amsmath,amsfonts}
\usepackage{algorithmic}
\usepackage{array}
\usepackage[caption=false,font=normalsize,labelfont=sf,textfont=sf]{subfig}
\usepackage{textcomp}
\usepackage{stfloats}
\usepackage{url}
\usepackage{verbatim}
\usepackage{graphicx}
\usepackage{orcidlink}
\usepackage{booktabs}
\usepackage{multirow}
\usepackage{amssymb}
\usepackage{etoolbox}
\makeatletter

\def\@IEEEBIOskipN{2\baselineskip}

\expandafter\patchcmd\csname\string\IEEEbiography\endcsname
  {\vskip \@IEEEBIOskipN plus 1fil minus 0\baselineskip}
  {\vskip \@IEEEBIOskipN}
  {}{}

\makeatother

\usepackage{pgfplots}
\pgfplotsset{compat=1.18}

\definecolor{covblue}{RGB}{55,112,180}
\definecolor{lenpink}{RGB}{232,0,125}
\definecolor{instorange}{RGB}{205,92,15}

\usepackage{xcolor}
\usepackage{tikz}

\definecolor{gtbg}{RGB}{218,228,246}
\definecolor{gtbar}{RGB}{135,170,220}

\definecolor{hrbg}{RGB}{224,239,215}
\definecolor{hrbar}{RGB}{165,205,140}

\definecolor{oursbg}{RGB}{252,222,202}
\definecolor{oursbar}{RGB}{245,175,120}

\def\BarH{0.70}

\def\BarTextSize{7.5}
\def\BarTextLine{8.5}

\def\TimeTextSize{6.0}
\def\TimeTextLine{8.0}

\def\LeftLabelSize{8.0}
\def\LeftLabelLine{9.0}

\newcommand{\BarTextFont}{\fontsize{\BarTextSize}{\BarTextLine}\selectfont}
\newcommand{\TimeTextFont}{\fontsize{\TimeTextSize}{\TimeTextLine}\selectfont}
\newcommand{\LeftLabelFont}{\fontsize{\LeftLabelSize}{\LeftLabelLine}\selectfont}

\providecommand{\timelinebg}[3]{}
\renewcommand{\timelinebg}[3]{%
    \pgfmathsetmacro{\yTop}{#1+\BarH}
    \pgfmathsetmacro{\yMid}{#1+0.5*\BarH}
    \fill[#3] (0,#1) rectangle (\D,\yTop);
    \node[font=\LeftLabelFont,anchor=east] at (-0.5,\yMid) {#2};
}

\providecommand{\timeseg}[5]{}
\renewcommand{\timeseg}[5]{%
    \pgfmathsetmacro{\yTop}{#2+\BarH}
    \pgfmathsetmacro{\yMid}{#2+0.5*\BarH}
    \fill[#4] (#1,#2) rectangle (#3,\yTop);

    \node[font=\BarTextFont,inner sep=0.5pt] at ({(#1+#3)/2},\yMid) {#5};

    \node[font=\TimeTextFont,anchor=north] at (#1,#2-0.04) {#1s};
    \node[font=\TimeTextFont,anchor=north] at (#3,#2-0.04) {#3s};
}

\usepackage{xcolor}
\usepackage{tikz}

\definecolor{GTBlue}{RGB}{150,160,255}
\definecolor{SeedGreen}{RGB}{170,210,170}
\definecolor{BeforeOrange}{RGB}{240,205,130}
\definecolor{AfterRedOrange}{RGB}{235,145,90}

\newcommand{\venue}[1]{$_{\mbox{\scriptsize #1}}$}

\def\BibTeX{{\rm B\kern-.05em{\sc i\kern-.025em b}\kern-.08em
    T\kern-.1667em\lower.7ex\hbox{E}\kern-.125emX}}
\usepackage{balance}
\begin{document}
\title{STAR: Structure-Aware Temporal Action Recovery for Point Supervised Temporal Action Localization}
\author{%
	Yawar~Rehman\,\orcidlink{0000-0001-9743-729X},
	Min~Young~Kim\,\orcidlink{0000-0001-7263-3403}, ~\IEEEmembership{Member,~IEEE}%
\thanks{Manuscript created May, 2026; This work was supported by the 2026 National University
Development Project funded by the Ministry of Education,
Republic of Korea.

(corresponding author: Min Young Kim)

Yawar Rehman is with Institute for Advanced Technology Convergence (Mobility), Kyungpook National University, Buk-gu, Daegu 41566, Republic of Korea (email: yawar@knu.ac.kr)

Min Young Kim is with School of Electronics Engineering, IT College, Kyungpook National University, Buk-gu, Daegu 41566, Republic of Korea (email: minykim@knu.ac.kr) }}

\markboth{IEEE TRANSACTIONS ON MULTIMEDIA, VOL.xx, NO.xx, 2026}%
{How to Use the IEEEtran \LaTeX \ Templates}
\IEEEaftertitletext{%
\begin{center}
{\Large\bfseries
This work was submitted to IEEE Transactions on Multimedia on September 01, 2026. And uploaded as preprint on personal website (https://yawar26.github.io/) while under review status on November 16, 2026. All rights reserved.
}
\end{center}
}

\maketitle
\footnotetext{This work has been submitted to the IEEE for possible publication. Copyright may be transferred without notice, after which this version may no longer be accessible.}

\begin{abstract}
Point supervised temporal action localization is becoming the need of time in this era of ever growing and non-stop video data. Existing method mainly rely on snippet level action response to generate pseudo labels. However, these labels often suffer from boundary ambiguity, over extended proposals, and weak temporal structure. To address these issues, we propose STAR, a structure-aware Temporal Action Recovery framework for point supervised temporal action localization. STAR works by enhancing the base S model with temporal difference interaction with attention module to capture forward and backward snippet changes, and introduces the learning  of start, core, and end roles to model the internal temporal structure of actions. In addition, STAR further improves the pseudo label generation through All point proposal ranking and Rolebank based structural re-ranking, which suppress neighbor crossing and structurally inconsistent proposals. Finally, temporal action detector to STAR feedback supervision transfers high quality detector prediction back to the STAR model through soft start, core, end, action, and background targets. This helps to further refine the pseudo labels. Extensive experiments on four standard benchmarks shows the effectiveness of STAR. By using I3D features, STAR achieves 67.2\% at AVG@0.1:0.7 on THUMOS'14, 58.1\% on GTEA, 74.6\% on BEOID, and 31.7\% at AVG@0.5:0.95 on ActivityNet-1.3, outperforming previous point supervised methods across multiple benchmarks.
\end{abstract}


\section{Introduction}
\IEEEPARstart{I}n the modern era, video has become a versatile way of transferring information. Lengthy sequences of data are recorded and saved from cameras, such as mobile phones, vehicle dashboard cameras, traffic monitoring cameras, and CCTV cameras. However, the information available in the recorded videos is limited, selective, and depends on the application. For example, insurance companies analyze recordings from dashcams in cases of accidents for compensation purposes, and CCTV camera recordings inside an industry are analyzed and used in cases of hazardous event information. Long stretches of video contain various background features, random motion activities, scene transitions, and temporally sparse actions. This makes the task of video understanding using deep learning models difficult. This highlights the need for research and better models in temporal action localization 
\begin{figure}[!t]
    \centering
    \includegraphics[
        width=\columnwidth,
        height=0.65\textheight,
        keepaspectratio
    ]{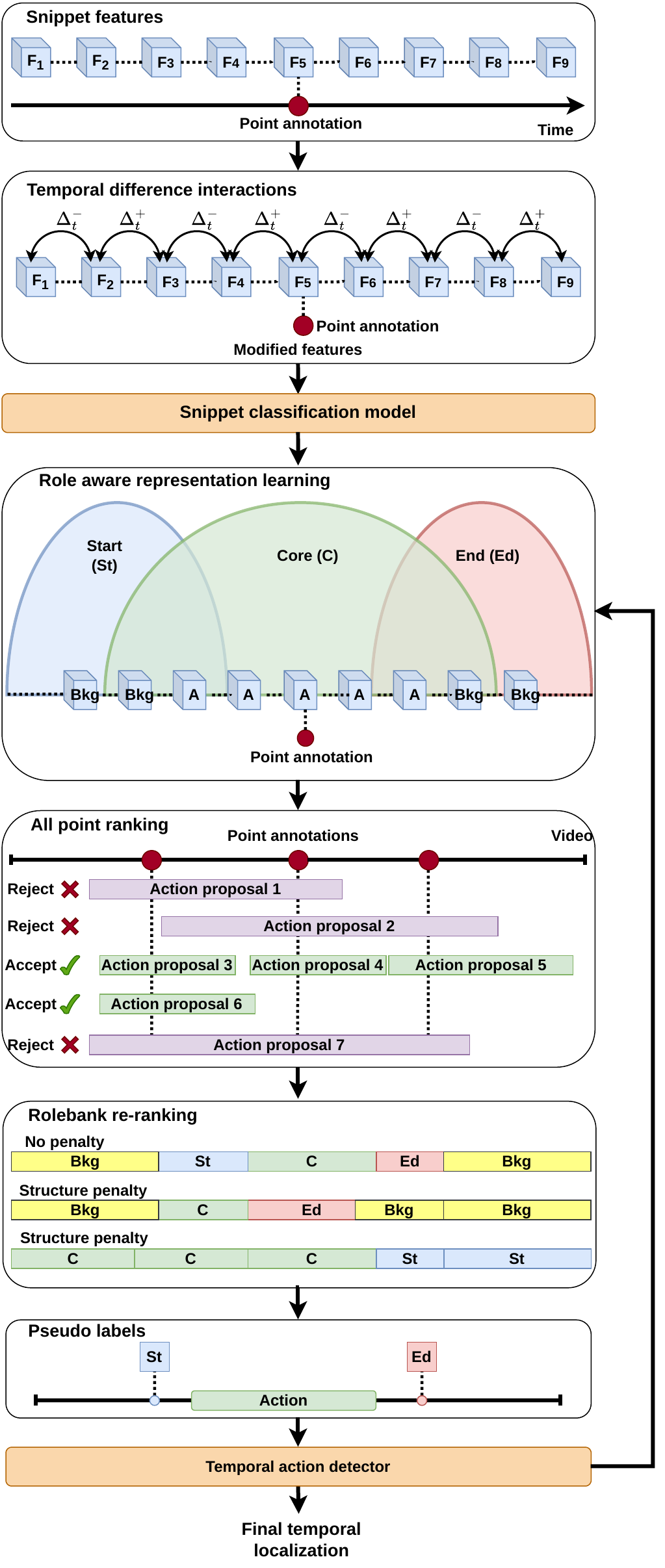}
    
    \caption{STAR at a glance: Structure-aware temporal action recovery (STAR). For snippet video features with point annotations, STAR model calculates temporal difference interaction between neighboring snippets. This enhance feature information is then forwarded to estimate CAS of action and background snippets. To recover the action structure from point supervision, STAR learns role aware representations for start, core, and end regions of action and background. Candidate action proposals are refined by All point ranking, which rejects action proposal that crosses neighboring point annotations. Action proposals are re-ranked with Rolebank re-ranking that penalizes action proposals whose structure is not similar to the action structure. Pseudo labels are then created and TAD is trained. Finally, TAD prediction are fed back to STAR to refine the role aware learning.}
    \label{fig:star_glance}
\end{figure}
(TAL) to aid video understanding. This includes determining the start and end time of the event of interest from long untrimmed videos.

Rapid progress has been made in fully supervised TAL \cite{shou2016temporal,zhao2017ssn,lin2017singleshot,
lin2018bsn,lin2019bmn,long2019gtan,xu2020gtad,
liu2020pbrnet}. These methods benefit from accurate temporal annotations in which the beginning and ending of every action instance are annotated and provided during training. However, this supervision requires substantial human effort to inspect a complete video and identify the multiple action boundaries inside a video. The cost and time increases when new video data is continuously collected and model has to be updated over time. This has motivated research towards weakly supervised and point supervised temporal action localization, where less expensive and faster forms of annotation are used to train the temporal action localization models. The point annotations for point supervised TAL give only partial information, however, it provides a more informative alternative than weakly supervised methods where a video is only identified to contain action type and no information is given about action location and number of action instances.

In point supervised TAL, the point annotations contain one temporal location belonging to each action instance during training, however, it does not indicate where the action starts and where it ends. A single annotated point does not provide a complete temporal structure of the action, therefore, recovering an entire action interval remains a problem.

In this work as shown in Figure~\ref{fig:star_glance}, we investigate this problem, as the main difficulty in point supervised temporal action localization is the lack of temporal structural information around the annotated point. An action instance contains temporal roles such as a starting scene, action core, and ending scene. Neighboring snippets contain temporal change information that can indicate when the visual representation moves from background to action or from action to background.

Based on this observation, we take the S model of HRPro \cite{zhang2024hrpro} as the starting core of our framework. However, its learning remained centered on the classification of discriminative snippet around annotated point. This is useful to indicate the presence of action but it does not explicitly informs about how the action develops temporally from beginning to center and towards the end. Therefore, we reformulate this core model into a Structure-Aware Temporal Action Recovery (STAR) network.

This structured representation of action using a single point annotations provide more reliable pseudo temporal annotations for training a TAD that leads to improved temporal localization performance on THUMOS'14, GTEA, BEOID, and ActivityNet-1.3 action datasets. The main contributions of this work are summarized as follows.

1. We propose Structure-Aware Temporal Action Recovery (STAR), which learns temporal structure of action instances using the dedicated start, core, and end prediction heads. By constructing the soft Gaussian targets from sparse point annotations, STAR transforms the point supervision into structured temporal supervision for action structure recovery. In addition, we propose a temporal difference interaction with an attention module to capture the relatively small motion changes in between snippets. We model the forward and backward changes between the neighboring snippets, which enables STAR to focus on motion classes that exhibit subtle motion differences from their backgrounds.

2. We introduce two action proposal ranking schemes, All point ranking and Rolebank re-ranking. All point scheme ranks the action proposals on the basis of confidence values as well as jointly evaluates the extent to which a proposal transgress into neighboring point annotations. It rejects the proposals whose temporal length transgress into neighboring point annotation. Furthermore, Rolebank re-ranks the proposals using start, core, and end role similarities and penalizes structurally unreliable candidate proposals. Together, the proposed ranking schemes bolster STAR to focus on reliable action proposals with reasonable temporal lengths and logical temporal structural continuity.

3. We propose feedback supervision from temporal action detector to STAR network. The detector proposals are converted into core, start, end, and background targets for refining the STAR learning using the robustness of temporal action detector.

4. Our propose method STAR network achieves state-of-the-art performance on the four standard benchmark point annotation based temporal action detection datasets. Our method achieves 67.2\% on THUMOS'14, 58.1\% on GTEA, 74.6\% on BEOID, and 31.7\% on ActivityNet-1.3 datasets using their corresponding evaluation metrics. 

\section{Related Work}
This section review the existing temporal action detection methods under fully and point supervised mechanisms and also focuses on Pseudo proposal refinement methods.
\subsection{Fully Supervised Temporal Action Localization}
Fully supervised temporal action localization has gradually moved from proposal generation and boundary regression towards anchor free and transformer based detectors. ActionFormer uses a multiscale temporal feature pyramid with local self attention to classify actions and regress their boundaries \cite{zhang2022actionformer}. TriDet improves the localization of action boundaries through relative boundary modeling \cite{shi2023tridet}, whereas DyFADet adapts temporal feature aggregation according to the duration and content of an action \cite{yang2024dyfadet}. Recently the Boundary Discretization and Reliable Classification Network has treated boundary estimation as a discretized prediction problem and has considered classification reliability during detection \cite{fang2025bdrc}. These methods show that accurate temporal intervals can be learned when complete start and end annotations are available. 

\subsection{Point Supervised Temporal Action Localization}
Point supervision reduces annotation effort by providing only one labeled temporal position inside each action instance. SFNet used sparse frame labels together with action and background modeling to extend supervision beyond the annotated frames \cite{ma2020sfnet}. Learning Action Completness from Points (LACP) learned action completeness from point annotations by generating dense pseudo labels and searching for more complete action sequences \cite{lee2021lacp}. Proposal based point level supervised temporal action localization methods (PTAL) generated flexible duration proposals containing broader temporal information \cite{yin2023proposal}. HRPro improved this process through hierarchical reliability propagation at snippet and instance levels \cite{zhang2024hrpro}. Xia \textit{et al.} realigned proposal confidence using temporal saliency information \cite{xia2024realigning}. Semantic Neighbor Guided Pseudo Label Refinement (SNPR) used temporal neighbors to generate soft pseudo labels and semantic neighbors to refine them \cite{li2024neighbor}. Stepwise Multi grained Boundary Detector (SMBD) introduced background anchors and stepwise boundary learning to reduce ambiguousness near the start and end of action \cite{liu2024smbd}. Query Reformation and Optimal Transport (QROT) used reformed queries and optimal transport to associate sparse points with complete action instances \cite{liu2025qrot}. Pseudo Label Refinement (PseR) focused on improving the quality of seed proposals through proposal propagation and a refinement network \cite{fang2026pser}. Although these methods have made point supervision more reliable, most of them mainly use the point as a seed for score propagation, proposal selection, or for boundary estimation. The ordered internal structure of an action is not learned as separate starting, central, and ending roles to give actions complete structural identity using point annotations.

\subsection{Temporal Structure and Pseudo Proposal Refinement}

SDANet used the differences in between snippets to guide attention in video level weakly supervised temporal action localization \cite{zhou2025sdanet}. On the other hand, our proposed method STAR measures both backward and forward changes around point supervised action instances and combines those changes with explicit temporal role learning. SMBD learns action boundaries in multiple steps \cite{liu2024smbd}, while the recent point to boundary method of Yang \textit{et al.} uses a Gaussian prior to generate dense action proposals from sparse points \cite{yang2026pointboundary}. A recent paper has further studied temporal consistency through action completion, action order, and action regularity tasks \cite{ma2026temporalconsistency}. However, they do not use the structured representation adopted in STAR, where start, core, and end regions are learned separately through soft Gaussian targets. The closest starting method with our proposed method STAR is HRPro \cite{zhang2024hrpro}. Its S model provides useful action and background evidence, but we propose to change the main learning objective from reliable snippet discrimination to structured temporal action recovery. The proposed All point ranking does not evaluate a proposal only from its confidence, it also checks its temporal extent with its neighboring annotated points. RoleBank re-ranking then checks whether the start, core, end, and background of a candidate agree with the learned role representations. PseR also improves pseudo labels before detector training \cite{fang2026pser}, however, our method STAR, improves the selection of action proposals based on their structure and additionally forms a feedback path from the trained detector to the structure recovery network.
\begin{figure*}[t]
    \centering

    \subfloat[]{
        \includegraphics[width=0.75\textwidth]{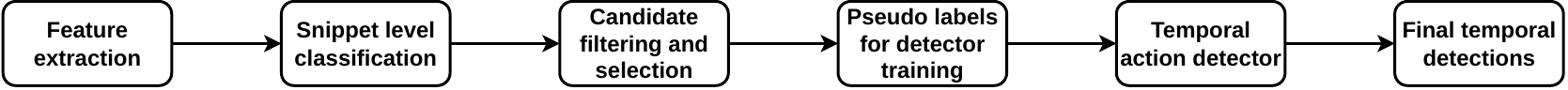}
        \label{fig:sota_block_diagram}
    }


    \subfloat[]{
        \includegraphics[width=0.75\textwidth]{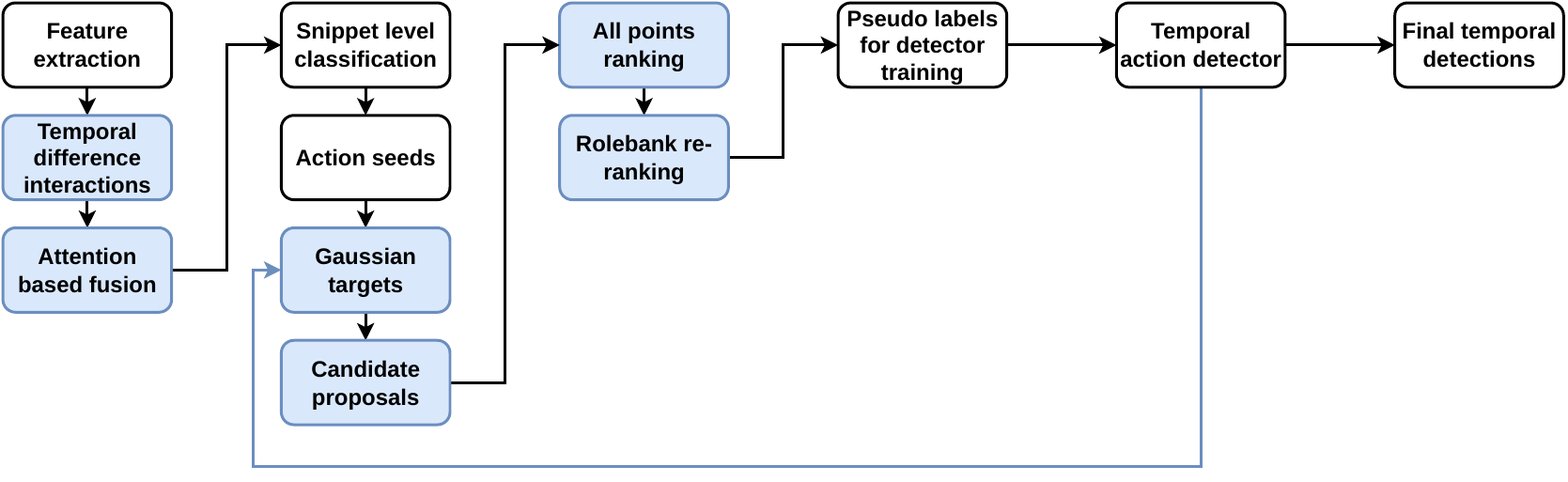}
        \label{fig:ours_block_diagram}
    }

    \caption{Comparison between the existing point supervised temporal action detection framework and our proposed STAR framework
    (a) SOTA framework 
    (b) Our proposed framework (blue highlighted blocks show our contribution)}
    \label{fig:block_diagrams}
\end{figure*}
\section{Proposed Method}
\label{sec:proposed_method}

We investigate the missing temporal structural information using the point annotations as an important factor for accurate action localization. Figure~\ref{fig:block_diagrams} highlights the contributions of the proposed method STAR, by comparing with the current SOTA methods.

\subsection{Problem Formulation}
\label{sec:problem_formulation}

Given an untrimmed video \(V\) is divided into \(T\) snippets and extracted its features \(F=\{F_t\}_{t=1}^{T}\), where \(F_t \in \mathbb{R}^{D}\) and \(D\) is the feature dimension. For point supervision networks, the training annotation for the \(i\)th action instance is represented as \(a_i=(t_i,c_i)\), where \(t_i\) is a single annotated temporal location belonging to the action and \(c_i\) represents its action category. Therefore, for a video containing \(N\) action instances, the available annotations are represented as \(\mathcal{A}=\{(t_i,c_i)\}_{i=1}^{N}\).

In contrast, complete temporal supervision would provide \(\mathcal{Y}=\{(s_i,e_i,c_i)\}_{i=1}^{N}\), where \(s_i\) and \(e_i\) are the actual starting and ending locations of the \(i\)th action instance, respectively. For point supervision, \(s_i\) and \(e_i\) are unavailable except for the \(t_i\) that is \(s_i<t_i<e_i\). Therefore, a large number of possible temporal intervals can satisfy the same point annotation, such as \([s_i^{(1)},e_i^{(1)}]\), \([s_i^{(2)},e_i^{(2)}]\), \(\ldots\), \([s_i^{(K)},e_i^{(K)}]\), although only some of these intervals can reliably represent the complete action.

A complete action contains temporally different regions that corresponds to its start, action core, and end, whereas the point annotation does not provides any information regarding these temporal roles. Hence, conventional point supervised learning mainly provides evidence that a feature snippet at the annotated location belongs to an action \(F_{t_i}\rightarrow\text{action}\), and does not provide information about the temporal role of surrounding snippets \(F_t\rightarrow\{\text{start},\text{core},\text{end}\}\). We refer this limitation as \emph{temporal structure ambiguity}. Our objective is to recover a structure-aware representation from the point annotations and further use it to obtain a reliable temporal proposal \(\hat{p}_i=(\hat{s}_i,\hat{e}_i,c_i)\)
\[
(t_i,c_i)
\rightarrow
\{\text{start},\text{core},\text{end}\}
\rightarrow
(\hat{s}_i,\hat{e}_i,c_i)
\]

This recovery is achieved by exploiting temporal changes between neighboring snippets, learning temporal role predictions, and evaluating whether candidate proposals are consistent with both the available point annotations and the learned action structure.

\subsection{Temporal Difference Interaction}
\label{sec:temporal_difference_interaction}
Local temporal variations inside the snippet features provide useful cues to recover the temporal structure of an action. Therefore, in this section, we introduce temporal difference interaction module to incorporate local temporal variation into the feature representation.

Given the input feature sequence \(F=\{F_t\}_{t=1}^{T}\), where \(F_t \in \mathbb{R}^{D}\) is the feature at temporal location \(t\), we first calculate the backward and forward temporal differences as follows.
\begin{equation}
\Delta_t^{-}=F_t-F_{t-1},
\qquad
\Delta_t^{+}=F_{t+1}-F_t
\label{eq:temporal_difference}
\end{equation}

\(\Delta_t^{-}\) describes the feature change from the previous snippet to the current snippet and \(\Delta_t^{+}\) describes the change from the current snippet to the following snippet. This is how we record the information about temporal variations.

After the information is recorded, instead of directly adding these differences to the original features, we use an attention mechanism to determine the relevant and irrelevant local temporal changes corresponding to the action sequence. The attention based temporal interaction feature is represented as follows.
\begin{equation}
M_t=\sum_{k}\alpha_{t,k}v_{t,k},
\qquad
\text{where}\qquad
\alpha_{t,k}=
\operatorname{softmax}\left(
\frac{q_t k_{t,k}^{T}}{\sqrt{d}}
\right)
\label{eq:temporal_attention}
\end{equation}

where \(q_t\) denotes the query representation of the current snippet, 
\(k_{t,k}\) and \(v_{t,k}\) denote the key and value representations of 
the neighboring temporal difference features, respectively, \(d\) is 
the feature dimension used for scaling, and \(\alpha_{t,k}\) represents 
the attention weight assigned to the \(k\)th neighboring temporal feature.  The resulting \(M_t\) represents the attention fused temporal change 
information at temporal location \(t\). The obtained temporal change representation is adaptively incorporated into the original snippet feature as follows.
\begin{equation}
\widetilde{F}_t =
\operatorname{LN}
\left(
F_t+\lambda_{\mathrm{td}}g_tM_t
\right)
\label{eq:feature_fusion}
\end{equation}

where \(g_t\) is a learnable gate that controls the contribution of temporal change information at each temporal location, \(\lambda_{\mathrm{td}}\) is a global scaling factor that controls the overall strength of temporal difference interaction, and \(\operatorname{LN}(\cdot)\) denotes layer normalization. Therefore, temporal changes are introduced selectively instead of being uniformly distributed across all snippets. The resulting feature sequence \(\widetilde{F}=\{\widetilde{F}_t\}_{t=1}^{T}\) is used for snippet level action and background prediction and temporal structure learning.

\begin{figure*}[t]
    \centering
    \includegraphics[width=\textwidth]{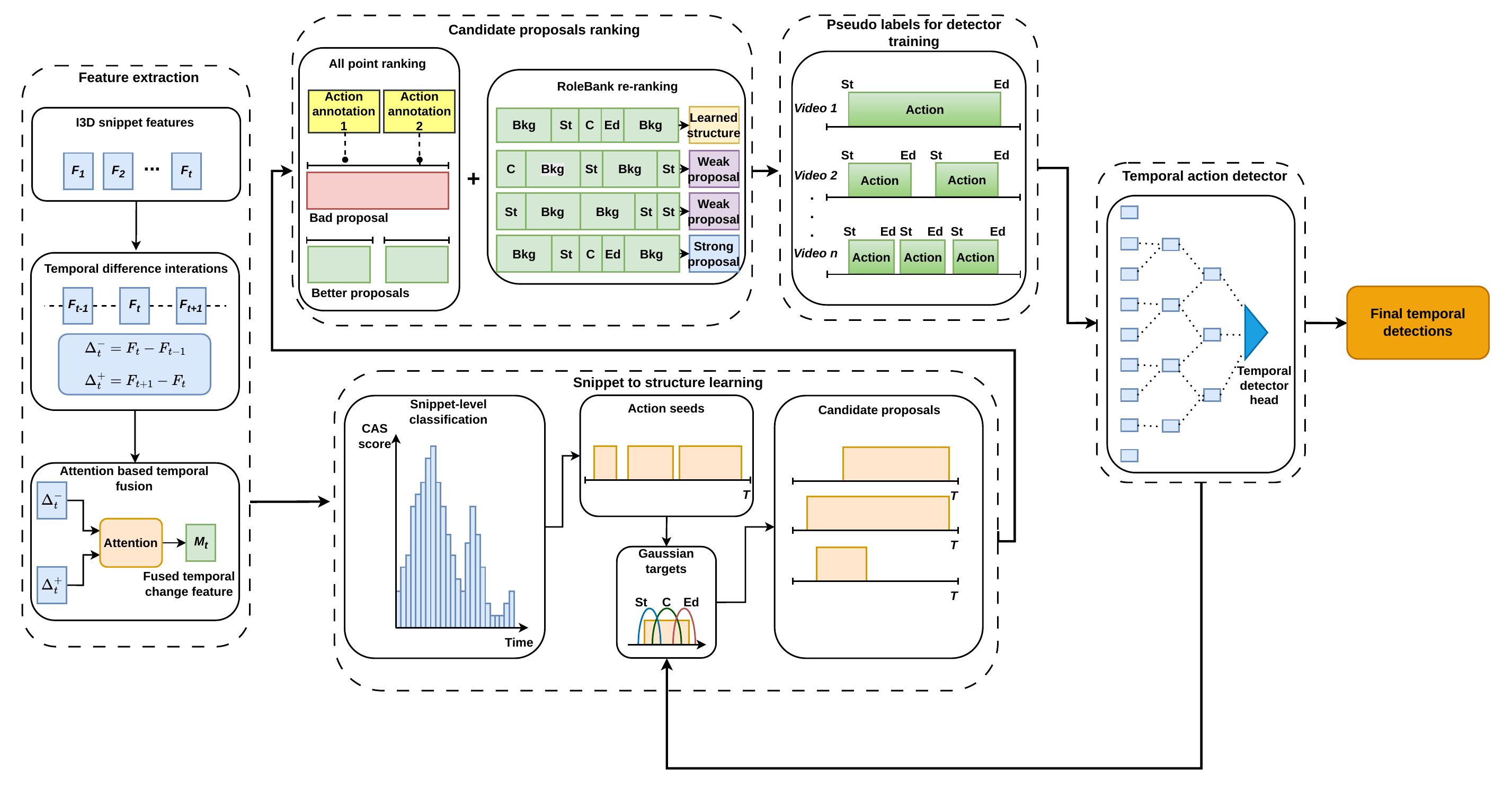}
    \caption{Overall architecture of the proposed STAR framework. STAR calculates the attention based temporal difference interations using the I3D snippets. The enhanced features are then forwarded to the S model for class activation sequence scores and action seeds. Using those seeds, STAR constructs soft Gaussian target  for temporal role learning. St, C, Ed, and Bkg represents Start, Core, End, and Background roles, respectively. Once candidate proposals are generated using the snippet level action responses those are refined through All point ranking and Rolebank re-ranking schemes to select the high quality action proposals. The selected action proposals are then converted to pseudo labels for the training of temporal action detector. Detector prediction are fed back to the STAR model as soft supervision for start, core, end, action, and background targets for further refinement of pseudo labels}    
    \label{fig:full_block_diagram}
\end{figure*}

\subsection{Structure-aware Temporal Role Learning}
\label{sec:structure_learning}

Once snippet features are enhanced with temporal interactions, those are given to the network encoder and snippet level classifier to estimate action and background probabilities as shown in Figure~\ref{fig:full_block_diagram}. The figure shows the full in detail process of our proposed method STAR. Instead of reinventing the wheel, we have used HRPro \cite{zhang2024hrpro} S model for snippet classification purpose. When \(C\) is the number of action classes and classifier produces a class activation sequence response (CAS), the action versus background snippets can be determined as follows.
\begin{equation}
P_{\mathrm{act}}=\operatorname{CAS}_{1:C},
\qquad
\text{and}\qquad
P_{\mathrm{bkg}}=\operatorname{CAS}_{C+1}
\label{eq:cas_probs}
\end{equation}

Background snippet response along with point annotations are used to expand the neighborhood contiguous action seeds around the annotated action locations. For the \(i\)th action instance, the resulting pseudo action seed can be represented as follows.
\begin{equation}
q_i =
[\widetilde{s}_i,\widetilde{e}_i],
\qquad
\widetilde{s}_i \leq t_i \leq \widetilde{e}_i
\label{eq:action_seed}
\end{equation}

Where \(t_i\) represents point annotation, \(\widetilde{s}_i\) and \(\widetilde{e}_i\) are the estimated start and end locations of neighborhood contiguous action seeds, respectively. These action seeds provide an approximation of the temporal extent for the corresponding action instance and are the initial source of temporal structural supervision.

The ground truth values for start and end boundaries are not available for point annotations, and assigning hard start and end targets would not be plausible to model uncertainty at extreme locations. Moreover, action seeds are estimated from weak supervision and are prone to temporal uncertainty as well. Therefore, we utilize and construct soft Gaussian targets to represent start, core, and end temporal roles to model uncertainty at the extremes of actions. 

For a temporal location \(t\), a Gaussian target centered at \(c\) with standard deviation \(\sigma\) is defined as follows.
\begin{equation}
G(t;c,\sigma)
=
\exp
\left[
-\frac{1}{2}
\left(
\frac{t-c}{\sigma}
\right)^2
\right]
\label{eq:gaussian}
\end{equation}

For each contiguous action seed, the start Gaussian is centered at the estimated seed start \(\widetilde{s}_i\), end Gaussian is centered at the estimated seed end \(\widetilde{e}_i\), and the core Gaussian is centered at the annotated point \(t_i\). The structural targets for the \(i\)th action instance are formulated as follows.
\begin{equation}
\begin{gathered}
G_{\mathrm{start}}^{i}(t)
=G(t;\widetilde{s}_i,\sigma_{\mathrm{edge}}^{i}),\;
G_{\mathrm{end}}^{i}(t)
=G(t;\widetilde{e}_i,\sigma_{\mathrm{edge}}^{i}),\\
G_{\mathrm{core}}^{i}(t)
=G(t;t_i,\sigma_{\mathrm{core}}^{i})
\end{gathered}
\label{eq:role_gaussians}
\end{equation}

The Gaussian formulation assigns the highest supervision around the temporal role i.e. start, core, or end roles, while gradually reduces its strength for neighboring snippets. This provides tolerance to the uncertainty present in the pseudo temporal structure. Furthermore,  instead of using a fixed Gaussian target length for every action instance, its width is adapted according to the temporal length of the action seeds.
\begin{equation}
L_i =
\widetilde{e}_i-\widetilde{s}_i+1
\label{eq:seed_length}
\end{equation}

Let Equation \eqref{eq:seed_length} represent the temporal length of \(i\)th action seed. The Gaussian widths for the action core and boundary regions can be calculated as follows.
\begin{equation}
\sigma_{\mathrm{core}}^{i}
=
\max\left(
\sigma_{\mathrm{core}}^{\min},
0.20L_i
\right),
\;
\sigma_{\mathrm{edge}}^{i}
=
\max\left(
\sigma_{\mathrm{edge}}^{\min},
0.08L_i
\right)
\label{eq:sigma_values}
\end{equation}

Where \(\sigma_{\mathrm{core}}^{\min}\) and \(\sigma_{\mathrm{edge}}^{\min}\) are the minimum allowed values for Gaussian widths for ensuring that the Gaussian functions never becomes too narrow. Consequently, with the help of Equation \eqref{eq:sigma_values}, longer action seeds receive wider structural perspective, and shorter action seeds get narrower structural perspective. This prevents the same temporal spread from being used on action instances having different durations.

Now, to learn these temporal roles i.e. start, core, and end, the temporally enhanced features are first transformed by the encoder as $Z =
\operatorname{Encoder}(\widetilde{F})$.

Where \(Z\) represents the encoded snippet feature representation. Our proposed method uses the encoded feature representations \(Z\) for training three different heads to understand the roles of start, core, and end inside the corresponding action sequences. The head responsible for learning core action role produces the class specific temporal prediction. It is defined as \(P_{\mathrm{core}}=\mathcal{S}\left(H_{\mathrm{core}}(Z)\right) \text{where}\; P_{\mathrm{core}}\in\mathbb{R}^{T\times C}\). The start and end heads are class agnostic and produce a temporal response for each snippet as \(P_{\mathrm{start}}=\mathcal{S}\left(H_{\mathrm{start}}(Z)\right) \text{and}\; P_{\mathrm{end}}=\mathcal{S}\left(H_{\mathrm{end}}(Z)\right)\), respectively.

Where \(P_{\mathrm{start}},P_{\mathrm{end}}\in\mathbb{R}^{T}\). \(T\) represents the number of temporal snippets inside the action video and \(C\) represents the number of action classes. Furthermore, \(H_{\mathrm{core}}\), \(H_{\mathrm{start}}\), and \(H_{\mathrm{end}}\) denotes the temporal role prediction heads and \(\mathcal{S}(\cdot)\) represents the sigmoid activation. Hence, the core prediction learns action class information and the start and end heads learn class independent temporal boundary information.

The seed selection procedure divides the temporal snippets into selected action and background seeds and unselected regions according to the CAS response and point annotations. Based on these selected action and background seed locations, we construct supervision masks. A value of one (1) assigned to the mask indicates a snippet selected for supervision and a value of zero (0) indicates an unselected snippet. These unselected snippets are excluded from the structural loss. Masks are then applied and used for Gaussian soft targets for start, core, and end snippets. Hence, we use masked binary cross entropy loss to ensure that only the temporal locations selected by the supervision masks contribute to the start, core, and end losses and the unselected snippets are ignored.
\begin{equation}
\mathcal{L}_{\mathrm{core}}
=
\operatorname{BCE}_{\mathrm{mask}}
\left(
P_{\mathrm{core}},
G_{\mathrm{core}},
M_{\mathrm{core}}
\right)
\label{eq:core_loss}
\end{equation}

Where \(P_{\mathrm{core}}\) is the predicted core role response, \(G_{\mathrm{core}}\) represents the core Gaussian soft target, \(M_{\mathrm{core}}\) is the core supervision mask, and \(\mathcal{L}_{\mathrm{core}}\) define the core loss function. Similarly, the start and end objectives are formulated as follows.
\begin{equation}
\mathcal{L}_{\mathrm{start}}
=
\operatorname{BCE}_{\mathrm{mask}}
\left(
P_{\mathrm{start}},
G_{\mathrm{start}},
M_{\mathrm{boundary}}
\right)
\label{eq:start_loss}
\end{equation}
\begin{equation}
\mathcal{L}_{\mathrm{end}}
=
\operatorname{BCE}_{\mathrm{mask}}
\left(
P_{\mathrm{end}},
G_{\mathrm{end}},
M_{\mathrm{boundary}}
\right)
\label{eq:end_loss}
\end{equation}

Where \(M_{\mathrm{boundary}}\) is the mask used for boundary supervision. The masks are derived from the selected action and background seed locations, and the structural loss is applied only where pseudo supervision is available.

During inference the start, core, and end heads predict snippet level probability maps from the incoming encoded snippet features. With the help of these maps, we generate action proposals and we name those as boundary aware action proposals. We then add boundary aware proposals to threshold based CAS proposal set. This produces richer candidate proposal pool before ranking stage. Thus, STAR learns and uses structural information associated with the action's start, core, and end regions as well as temporal transition patterns to generate boundary aware action proposals and to recover the action structure.

\subsection{Candidate Proposal Ranking}
\label{sec:candidate_ranking}
The learned snippet action responses are used to generate the candidate temporal proposals. We investigate the two main issues in the generated temporal proposals. First, the point annotations does not contains start and end of action location information, therefore some of the generated candidate temporal proposals cross neighboring annotation action instances.
Secondly, some of the candidate temporal proposals with reasonable action confidence values had inconsistent temporal structure i.e. structure not similar to the start, core, and end structure as learned by the proposed STAR method. Therefore, we proposed two staged candidate proposal ranking strategy (1) All point ranking and (2) Rolebank re-ranking. Both are discussed as follows.

\subsubsection{All Point Ranking}
\label{sec:all_points}
All Point ranking evaluates each candidate proposal by constraining it with the neighboring action annotation points. For a candidate proposal \(p=[p_s,p_e]\), with an annotated point, its start and end locations are constrained by the temporally adjacent action point annotations as given below.
\begin{equation}
p_s > t_{\mathrm{prev}} \qquad and
\qquad
p_e < t_{\mathrm{next}}
\label{eq:neighbor_constraint}
\end{equation}

Where \(t_{prev}\) and \(t_{next}\) represents previous and next neighboring point annotations, respectively and \(p_s\) and \(p_e\) represents the start and end of the candidate proposal inside the same video. Those candidate proposals who fail to follow this constraint are removed from the candidate proposal pool. In case, if no valid candidate exists of a certain annotation point then the proposed All point ranking method restores back to the removed proposal. This fall back is useful and helps to avoid missing the annotation proposal completely.

This All point ranking is important when there are multiple action instances inside the same video. A candidate proposal having an annotation point should not be allowed to transgress inside another action instance annotation point. Hence, All points ranking method suppresses all such neighbor crossing candidate proposals and allows the proposals that satisfy the sparse point supervision.

\subsubsection{Rolebank re-Ranking}
\label{sec:all_points}
After All Points ranking, the remaining candidate proposals are further evaluated and re-ranked according to their internal temporal structure. We first define high quality action candidates as those candidates that have a good classification confidence as well as a consistent organization of start, core, and end regions. 

Then for each candidate proposal, the encoded feature vectors are extracted from start, core, end, and background regions. These local regional feature representations are compared with the role aware representations inside the Rolebank to evaluate whether the proposal exhibits the expected temporal organization. The Rolebank score for a candidate proposal \(p\) is formulated as follows.
\begin{equation}
S_{\mathrm{RB}}(p)
=
s_p
+
\alpha R(p)
+
\beta_{\mathrm{out}}B_{\mathrm{out}}(p)
-
\beta_{\mathrm{in}}B_{\mathrm{in}}(p)
-
P_{\mathrm{len}}(p)
\label{eq:rolebank}
\end{equation}

Where \(R(p)\) measures the structural consistency of the encoded proposal features with start, core, and end role representations. \(B_{\mathrm{out}}(p)\) measures the consistency of the feature representation outside the proposal with the background role, \(B_{\mathrm{in}}(p)\) penalizes if there are background like snippet features occurring inside the candidate proposal. \(P_{\mathrm{len}}(p)\) penalizes undesirable proposal lengths. \(P_{\mathrm{len}}(p)\) uses class specific length prior calculated from the proposals obtained from All point ranking step for penalizing the undesirable proposal lengths. The structural role consistency \(R(p)\) is calculated as follows.
\begin{equation}
R(p)
=
\operatorname{mean}
\left(
\operatorname{sim}_{\mathrm{start}}(p),
\operatorname{sim}_{\mathrm{core}}(p),
\operatorname{sim}_{\mathrm{end}}(p)
\right)
\label{eq:role_similarity}
\end{equation}

Where \(\operatorname{sim}_{\mathrm{start}}(p)\), \(\operatorname{sim}_{\mathrm{core}}(p)\), and \(\operatorname{sim}_{\mathrm{end}}(p)\) are the similarities between the encoded features extracted from the start, core, and end portions of proposal \(p\) and their respective role aware representations in Rolebank. Rolebank is built from boundary aware action proposals by extracting encoded snippet features from their start, core, and end regions. Thus, the top candidate proposal is not selected solely as per its classification confidence, its structural role is also taken into account.

The proposed ranking methods first restrict the candidate proposal sliding into the neighboring action annotation points. And then uses the structural information to further re-rank the candidate proposals. This filters out most of the unwanted candidate proposals and we are left with only strong candidates that contains a single annotation point and duly follows structural start, core, and end roles. The highest ranked proposals are then selected as pseudo labels for training the TAD. 

\subsection{Temporal Action Detector's Feedback}
\label{sec:tadf}

The pseudo labels generated by our proposed method STAR are forwarded to temporal action detector (TAD) for learning the temporal action localization. TAD due to its boundary localization objective function, inherently possesses a stronger temporal localization capability as it learns to refine the start and end boundaries of action instances. This reduces the localization error induced by the weakly supervised pseudo boundaries. Once the training of TAD is complete, we generate additional teacher targets for action start, core, end, and background regions. These derived targets from TAD are given to the STAR network for fixing the boundary error.

The generated teacher targets for action start, core, end, and background are represented as \(T_{\mathrm{fb\_start}}\), \(T_{\mathrm{fb\_core}}\),  \(T_{\mathrm{fb\_end}}\), \(T_{\mathrm{action}}\), and \(T_{\mathrm{bkg}}\), respectively. The structural prediction heads are refined as follows.
\begin{equation}
\begin{aligned}
\mathcal{L}_{\mathrm{fb\_core}}
&=
\operatorname{BCE}
\left(
P_{\mathrm{core}},
T_{\mathrm{fb\_core}}
\right),\\
\mathcal{L}_{\mathrm{fb\_start}}
&=
\operatorname{BCE}
\left(
P_{\mathrm{start}},
T_{\mathrm{fb\_start}}
\right),\\
\mathcal{L}_{\mathrm{fb\_end}}
&=
\operatorname{BCE}
\left(
P_{\mathrm{end}},
T_{\mathrm{fb\_end}}
\right)
\end{aligned}
\label{eq:feedback_losses}
\end{equation}

Similarly, the TAD generated temporal regions are used to refine the action and background responses as follows.
\begin{equation}
\begin{aligned}
\mathcal{L}_{\mathrm{fb\_action}}
&=
\operatorname{BCE}
\left(
P_{\mathrm{act}},
T_{\mathrm{action}}
\right),\\
\mathcal{L}_{\mathrm{fb\_bkg}}
&=
\operatorname{BCE}
\left(
P_{\mathrm{bkg}},
T_{\mathrm{bkg}}
\right)
\end{aligned}
\label{eq:feedback_action_bkg}
\end{equation}

This establishes a feedback path between the proposed method STAR and fully trained temporal action detector. Using the point annotations, STAR initially recovers temporal structure of actions and generate pseudo labels for the TAD training. Then the TAD give its learned localization information to the STAR. Therefore, the stronger temporal localization knowledge of TAD has been used as a supervision for refining the action structure recovered from point annotations.

\section{Experiments}
\label{sec:exp}

\subsection{Datasets, Evaluation Metrics, and Experimental Setup}
\label{sec:exp_setup}
\textit{Datasets:} We evaluated the proposed STAR method on four point supervised temporal action localization standard datasets named as THUMOS'14, GTEA, BEOID, and ActivityNet-1.3. For fair comparison we use the same point annotations and dataset splits as available at \cite{hrprogithub}. THUMOS'14 dataset contains 20 action classes with 200 training videos and 213 testing videos. GTEA dataset contains 28 videos from 7 kitchen activities. BEOID dataset contains 58 videos from 30 action classes. ActivityNet-1.3 10,024 training video, 4,926 validation and 5,044 test video with 200 action classes. It is the biggest dataset among all the datasets we experimented and tested our proposed method STAR.

\textit{Evaluation Metrics:} Mean Average Precision (mAP) on multiple intersections over unions (IoU) is the standard way to evaluation networks trained using point annotation for temporal action localizations. For THUMOS'14, GTEA, and BEOID datasets we use standard mAP of IoUs in the range [0.1:0.1:0.7], however for ActivityNet-1.3 we follow the standard protocol of reporting accuracy and mAP of tIoUs in the range [0.5:0.05:0.95].

\textit{Experimental Setup:} 
For the THUMOS'14 I3D features before the feedback, we set learning rate as 0.0001, weight decay as 0.0005, and evaluation after every 50 iterations. With these parameters, the training of the proposed STAR model continues for 4000 iterations. Then after the information is relayed back to STAR model to learn the targets for start, core, end, and background in the feedback, the STAR model is trained for 2000 iterations with learning rate as 0.00005 and weight decay as 0.0005. We follow the same settings for the THUMOS'14 VideoMAE features \cite{videomaev2_tad}. The detailed experimental settings are summarized in Table~\ref{tab:training_settings}.
\begin{table}[!ht]
\centering
\caption{Train settings of STAR across different datasets}
\label{tab:training_settings}
\renewcommand{\arraystretch}{1.12}
\setlength{\tabcolsep}{2.8pt}
\scriptsize
\begin{tabular}{l|l|cccc}
\hline\hline
Dataset
& Stage
& LR
& WD
& Iter.
& Eval. \\
\hline

\multirow{2}{*}{\shortstack[l]{THUMOS'14}}
& Before FB
& $1\times10^{-4}$
& $5\times10^{-4}$
& 4000
& 50 \\
& After FB
& $5\times10^{-5}$
& $5\times10^{-4}$
& 2000
& 50 \\
\hline

\multirow{2}{*}{\shortstack[l]{GTEA}}
& Before FB
& $1\times10^{-4}$
& $1\times10^{-3}$
& 1000
& 50 \\
& After FB
& $1\times10^{-4}$
& $1\times10^{-2}$
& 1000
& 50 \\
\hline

\multirow{2}{*}{\shortstack[l]{BEOID}}
& Before FB
& $1\times10^{-4}$
& $1\times10^{-3}$
& 1000
& 50 \\
& After FB
& $5\times10^{-5}$
& $1\times10^{-3}$
& 1000
& 50 \\
\hline

\multirow{2}{*}{ActivityNet-1.3}
& Before FB
& $1\times10^{-4}$
& $5\times10^{-4}$
& 30000
& 1000 \\
& After FB
& $1\times10^{-3}$
& $5\times10^{-2}$
& 30000
& 1000 \\
\hline\hline
\end{tabular}
\end{table}
\subsection{Comparison with State-of-the-Art methods}
\label{sec:sota_compare}
In this section, we compare the proposed STAR method with state-of-the-art based on point based weakly supervised temporal action localization methods. Full frame level supervised methods are also reported for reference. 

The obtained results on THUMOS'14 are shown in Table~\ref{tab:thumos14_detection}. Our proposed STAR method achieves the best performance among point level weakly supervised methods, obtaining 67.2\% at AVG@0.1:0.7, 60.6\% at AVG@0.3:0.7, and 76.9\% at AVG@0.1:0.5. As compared with the recent strongest point supervised method known as PseR, STAR improves AVG@0.1:0.7 by 2.4 percentage points and AVG@0.3:0.7 by 3.7 percentage points. The improvement is especially clear at stricter IoU thresholds, where STAR improves mAP@0.5, mAP@0.6, and mAP@0.7 from 60.1\%, 45.4\%, and 29.6\% to 64.3\%, 51.6\%, and 34.6\%, respectively. This indicates that STAR does not only detect action presence but also produces temporally more accurate action boundaries.

Compared with HRPro, which is our base point supervised framework, STAR improves from 60.3\% to 67.2\%. The gain mainly comes from the proposed structure aware role learning. Although STAR uses only point annotations, its performance is close to fully supervised ActionFormer, hence, reducing the gap between weak and full supervision.

Table~\ref{tab:gtea_beoid_activitynet} reports the results on GTEA, BEOID, and ActivityNet-1.3. On GTEA, our proposed method STAR achieves 58.1\%, outperforming the previous best point supervised method PseR by 5.8 percent. On BEOID, STAR achieves 74.6\%, improving over PseR by 5.7 percent. The improvement is particularly strong at high IoU thresholds. On ActivityNet-1.3, our method STAR achieves 31.7\% at AVG@0.5:0.95, surpassing the previous best point supervised method PseR by 1.1 percent. STAR also improves mAP@0.5 and mAP@0.75 to 51.0\% and 31.4\%, respectively. Our proposed method STAR consistently improves the average mAP. This demonstrates the better generalization capability of the proposed method across dense short action dataset and large scale long action dataset.
\begin{table*}[t]
\centering
\caption{Comparison of Full, Weak, and Point supervised detection performance on THUMOS'14}
\label{tab:thumos14_detection}
\renewcommand{\arraystretch}{1.08}
\setlength{\tabcolsep}{3.05pt}
\resizebox{0.7\textwidth}{!}{
\begin{tabular}{c|l|ccccccc|ccc}
\hline\hline
\multirow{2}{*}{Supervision}
& \multicolumn{1}{c|}{\multirow{2}{*}{Method}}
& \multicolumn{7}{c|}{mAP@IoU(\%)}
& \multirow{2}{*}{\shortstack{AVG\\(0.1:0.5)}}
& \multirow{2}{*}{\shortstack{AVG\\(0.3:0.7)}}
& \multirow{2}{*}{\shortstack{AVG\\(0.1:0.7)}} \\
&
& 0.1 & 0.2 & 0.3 & 0.4 & 0.5 & 0.6 & 0.7
& & & \\
\hline

\multirow{3}{*}{\shortstack{Frame-level\\(Full)}}
& AFSD\venue{CVPR'21}~\cite{lin2021afsd}
& - & - & 67.3 & 62.4 & 55.5 & 43.7 & 31.1 & - & 52.0 & - \\
& ActionFormer\venue{ECCV'22}~\cite{zhang2022actionformer}
& - & - & 82.1 & 77.8 & 71.0 & 59.4 & 43.9 & - & 66.8 & - \\
& DyFADet\venue{ECCV'24}~\cite{yang2024dyfadet}
& - & - & \textbf{84.0} & \textbf{80.1} & \textbf{72.7} & \textbf{61.1} & \textbf{47.9} & - & \textbf{69.2} & - \\
\hline

\multirow{11}{*}{\shortstack{Video-level\\(Weak)}}
& KT-MGFN\venue{TMM'20}~\cite{su2021ktmgfn}
& 56.5 & 51.3 & 42.2 & 33.8 & 25.6 & 17.0 & 7.6 & 41.9 & 25.2 & 33.4 \\
& Li \textit{et al.}\venue{TMM'22}~\cite{li2023superframe}
& - & - & - & 41.7 & 32.9 & 22.3 & 12.5 & - & - & - \\
& MDA\venue{TMM'22}~\cite{chen2023mda}
& 69.7 & 63.1 & 55.2 & 46.6 & 35.6 & 25.0 & 14.4 & 54.0 & 35.4 & 44.2 \\
& Li \textit{et al.}\venue{CVPR'23}~\cite{li2023text}
& - & - & 56.2 & 47.8 & 39.3 & 27.5 & 15.2 & - & 37.2 & - \\
& P-MIL\venue{CVPR'23}~\cite{ren2023pmil}
& 71.8 & 67.5 & 58.9 & 49.0 & 40.0 & 27.1 & 15.1 & 57.4 & 38.0 & 47.0 \\
& PivoTAL\venue{CVPR'23}~\cite{rizve2023pivotal}
& 74.1 & 69.6 & \textbf{61.7} & 52.1 & 42.8 & \textbf{30.6} & 16.7 & 60.1 & 40.8 & 49.6 \\
& Zhou \textit{et al.}\venue{CVPR'23}~\cite{zhou2023bridging}
& 74.0 & 69.4 & 60.7 & 51.8 & 42.7 & 26.2 & 13.1 & 59.7 & 38.9 & 48.3 \\
& F3-Net\venue{TMM'23}~\cite{moniruzzaman2024f3net}
& 72.0 & 66.1 & 56.5 & 48.2 & 38.9 & - & - & 56.3 & - & - \\
& ISSF\venue{AAAI'24}~\cite{yun2024issf}
& 72.4 & 66.9 & 58.4 & 49.7 & 41.8 & 25.5 & 12.8 & 57.8 & 37.6 & 46.8 \\
& Li \textit{et al.}\venue{MM'24}~\cite{li2024dualconfounding}
& 73.0 & 69.2 & 59.6 & 50.2 & 41.8 & 28.4 & 14.9 & 58.8 & 39.0 & 48.2 \\
& PVLR\venue{MM'24}~\cite{lim2024pvlr}
& \textbf{74.9} & \textbf{69.9} & 61.4 & \textbf{53.1} & \textbf{45.1} & 30.5 & \textbf{17.1} & \textbf{60.9} & \textbf{41.4} & \textbf{50.3} \\
\hline

\multirow{11}{*}{\shortstack{Point-level\\(Weak)}}
& SF-Net\venue{ECCV'20}~\cite{ma2020sfnet}
& 68.3 & 62.3 & 52.8 & 42.2 & 30.5 & 20.6 & 12.0 & 51.2 & 31.6 & 41.2 \\
& Ju \textit{et al.}\venue{ICCV'21}~\cite{ju2021divide}
& 72.3 & 64.7 & 58.2 & 47.1 & 35.9 & 23.0 & 12.8 & 55.6 & 35.4 & 44.9 \\
& LACP\venue{ICCV'21}~\cite{lee2021lacp}
& 75.7 & 71.4 & 64.6 & 56.5 & 45.3 & 34.5 & 21.8 & 62.7 & 44.5 & 52.8 \\
& CRRC-Net\venue{TIP'22}~\cite{fu2022crrcnet}
& 77.8 & 73.5 & 67.1 & 57.9 & 46.6 & 33.7 & 19.8 & 64.6 & 45.1 & 53.8 \\
& LPR\venue{MMSJ'23}~\cite{fang2023lpr}
& 85.3 & 82.5 & 76.6 & 67.8 & 55.3 & 38.5 & 21.7 & 73.5 & 52.0 & 61.1 \\
& Yin \textit{et al.}\venue{BMVC'23}~\cite{yin2023proposal}
& 77.1 & 72.6 & 65.9 & 54.4 & 44.9 & 33.1 & 20.2 & 63.3 & 43.9 & 52.6 \\
& HRPro\venue{AAAI'24}~\cite{zhang2024hrpro}
& \textbf{85.6} & 81.6 & 74.3 & 64.3 & 52.2 & 39.8 & 24.8 & 71.6 & 51.1 & 60.3 \\
& TSPNet\venue{CVPR'24}~\cite{xia2024realigning}
& 82.3 & 77.6 & 70.1 & 60.0 & 49.4 & 37.6 & 24.5 & 67.9 & 48.3 & 57.4 \\
& SMBD\venue{ECCV'24}~\cite{liu2024smbd}
& - & - & 66.0 & 57.9 & 47.0 & 36.0 & 22.0 & 64.2 & 45.7 & - \\
& PseR\venue{TMM'26}~\cite{fang2026pser}
& 85.5 & \textbf{83.2} & 78.7 & 70.9 & 60.1 & 45.4 & 29.6 & 75.7 & 56.9 & 64.8 \\
& STAR (Ours)
& 84.9
& 82.8
& \textbf{79.5}
& \textbf{72.9}
& \textbf{64.3}
& \textbf{51.6}
& \textbf{34.6}
& \textbf{76.9}$^{\uparrow1.2}$
& \textbf{60.6}$^{\uparrow3.7}$
& \textbf{67.2}$^{\uparrow2.4}$ \\
\hline\hline
\end{tabular}
}
\end{table*}
\begin{table*}[t]
\centering
\caption{Comparison of detection performance on GTEA, BEOID, and ActivityNet-1.3}
\label{tab:gtea_beoid_activitynet}
\renewcommand{\arraystretch}{1.08}
\setlength{\tabcolsep}{3.8pt}
\resizebox{0.85\textwidth}{!}{
\begin{tabular}{c|l|ccccc|ccccc|cccc}
\hline\hline
\multirow{3}{*}{Supervision}
& \multirow{3}{*}{Method}
& \multicolumn{5}{c|}{GTEA}
& \multicolumn{5}{c|}{BEOID}
& \multicolumn{4}{c}{ActivityNet-1.3} \\
&
& \multicolumn{5}{c|}{mAP@tIoU (\%)}
& \multicolumn{5}{c|}{mAP@tIoU (\%)}
& \multicolumn{4}{c}{mAP@tIoU (\%)} \\
&
& 0.1 & 0.3 & 0.5 & 0.7 & {\scriptsize AVG$[0.1:0.7]$}
& 0.1 & 0.3 & 0.5 & 0.7 & {\scriptsize AVG$[0.1:0.7]$}
& 0.5 & 0.75 & 0.95 & {\scriptsize AVG$[0.5:0.95]$} \\
\hline

\multirow{4}{*}{\shortstack{Video-level\\(weak)}}
& PivoTAL\venue{CVPR'23}~\cite{rizve2023pivotal}
& - & - & - & - & -
& - & - & - & - & -
& \textbf{45.1} & 28.2 & 5.0 & 28.1 \\
& Zhou \textit{et al.}\venue{CVPR'23}~\cite{zhou2023bridging}
& - & - & - & - & -
& - & - & - & - & -
& 43.4 & \textbf{28.8} & \textbf{9.9} & \textbf{28.8} \\
& ISSF\venue{AAAI'24}~\cite{yun2024issf}
& - & - & - & - & -
& - & - & - & - & -
& 39.4 & 25.8 & 6.4 & 25.8 \\
& PVLR\venue{MM'24}~\cite{lim2024pvlr}
& - & - & - & - & -
& - & - & - & - & -
& 43.6 & 27.4 & 6.5 & 27.4 \\
\hline

\multirow{10}{*}{\shortstack{Point-level\\(weak)}}
& SF-Net\venue{ECCV'20}~\cite{ma2020sfnet}
& 58.0 & 37.9 & 19.3 & 11.9 & 31.0
& 62.9 & 40.6 & 16.7 & 3.5 & 30.9
& - & - & - & - \\
& Ju \textit{et al.}\venue{ICCV'21}~\cite{ju2021divide}
& 59.7 & 38.3 & 21.9 & 18.1 & 33.7
& 63.2 & 46.8 & 20.9 & 5.8 & 34.9
& - & - & - & - \\
& LACP\venue{ICCV'21}~\cite{lee2021lacp}
& 63.9 & 55.7 & 33.9 & 20.8 & 43.5
& 76.9 & 61.4 & 42.7 & 25.1 & 51.8
& 40.4 & 24.6 & 5.7 & 25.1 \\
& CRRC-Net\venue{TIP'22}~\cite{fu2022crrcnet}
& - & - & - & - & -
& - & - & - & - & -
& 39.8 & 24.1 & 5.9 & 24.0 \\
& Yin \textit{et al.}\venue{BMVC'23}~\cite{yin2023proposal}
& 63.1 & 52.1 & 37.3 & 22.2 & 45.1
& 78.2 & 65.3 & 45.1 & 26.6 & 54.2
& 48.3 & 27.8 & 7.0 & 29.1 \\
& HRPro\venue{AAAI'24}~\cite{zhang2024hrpro}
& 72.6 & 61.1 & 37.3 & 17.5 & 47.3
& 78.5 & 72.1 & 55.3 & 26.1 & 59.4
& 42.8 & 27.2 & \textbf{8.0} & 27.1 \\
& TSPNet\venue{CVPR'24}~\cite{xia2024realigning}
& 74.6 & 60.9 & 39.5 & 16.6 & 49.0
& 83.8 & 73.0 & 51.1 & 23.8 & 59.6
& - & - & - & - \\
& SMBD\venue{ECCV'24}~\cite{liu2024smbd}
& 75.0 & 61.3 & 41.1 & 14.2 & 47.4
& 78.2 & 71.0 & 52.5 & 25.2 & 57.4
& - & - & - & - \\
& PseR\venue{TMM'26}~\cite{fang2026pser}
& 72.7 & 67.7 & 42.9 & 19.9 & 52.3
& 88.9 & 83.9 & 66.1 & 28.9 & 68.9
& 49.5 & 29.6 & 7.8 & 30.6 \\
& STAR (Ours)
& \textbf{80.5} & \textbf{75.0} & \textbf{50.6} & \textbf{23.9} & $\mathbf{58.1}^{\uparrow 5.8}$
& \textbf{90.9} & \textbf{86.7} & \textbf{72.1} & \textbf{43.8} & $\mathbf{74.6}^{\uparrow 5.7}$
& \textbf{51.0} & \textbf{31.4} & 4.8 & $\mathbf{31.7}^{\uparrow 1.1}$ \\

\hline\hline
\end{tabular}
}
\end{table*}

\begin{table}[t]
\centering
\caption{Ablation study of the proposed STAR method with I3D features}
\label{tab:ablation_study}
\renewcommand{\arraystretch}{1.08}
\setlength{\tabcolsep}{4pt}
\footnotesize
\resizebox{0.75\columnwidth}{!}{%
\begin{tabular}{l|cccc}
\hline\hline

Baseline
& \checkmark & \checkmark & \checkmark & \checkmark \\

Role learning and ranking
& & \checkmark & \checkmark & \checkmark \\

Temporal difference features
& & & \checkmark & \checkmark \\

Feedback
& & & & \checkmark \\
\hline

AVG (0.1:0.7)
& 62.2 & 65.9 & 66.9 & 67.2 \\

\hline\hline
\end{tabular}
}
\end{table}

\subsection{Ablation Study}
\label{sec:ablation}
We conducted ablation experiments on THUMOS'14 to analyze the contribution of each proposed component. We followed the previous works for baseline (PseR \cite{fang2026pser}) and used HRPro as point based weakly supervised temporal action detection, ActionFormer as temporal action detector and seed proposals as pseudo labels. We then added the proposed role learning and ranking modules, then introduced temporal difference features, and finally used the feedback supervision. Table~\ref{tab:ablation_study} shows the results using I3D features. The baseline obtained 62.2\%. After adding role learning and ranking, the performance increases to 65.9\%, showing a gain of 3.7 percentage points. This indicates that modeling start, core, and end roles and using them for proposal ranking improves the quality of generated pseudo labels. Adding temporal difference features further improves the result to 66.9\%. This demonstrates that temporal change information is useful for action localization. Finally, the feedback stage increases the performance to 67.2\%, giving the best result with I3D features. We also give the addition ablation studies with VideoMAE features in supplementary material Section I(A).
\subsection{Hyperparameter Analysis}
\label{sec:hype}

Table~\ref{tab:hyperparameter_ablation} shows the analysis on the effect of feedback supervision weights. We first vary action and background weights while keeping the boundary weights frozen. We increased the background weight from 0.05 to 0.20, which consistently improved the final action detector performance as gave 67.24\%. However, as we increased the action weights from 0.05 to 0.20, the performance reduced. It means that background region supervision is important than the action region supervision and it also overconstrains the model.

We then keep the action and background weights constant and vary the boundary weights i.e. core, start and end weights. We got the best results when core, start, and end is 0.20, 0.08, 0.08, respectively.
\begin{table}[t]
\centering
\caption{Hyperparameter ablation of action, background, and boundary weights in the feedback supervision}
\label{tab:hyperparameter_ablation}
\renewcommand{\arraystretch}{1.08}
\setlength{\tabcolsep}{6pt}
\resizebox{0.80\columnwidth}{!}{%
\begin{tabular}{l|cc|ccc|c}
\hline\hline
\multicolumn{1}{c|}{Method}
& \multicolumn{1}{c}{Act}
& \multicolumn{1}{c|}{Bkg}
& \multicolumn{1}{c}{Core}
& \multicolumn{1}{c}{Start}
& \multicolumn{1}{c|}{End}
& \multicolumn{1}{c}{AVG (0.1:0.7)} \\
\hline

Ours & 0.05 & 0.05 & 0.20 & 0.08 & 0.08 & 66.13 \\
Ours & 0.05 & 0.10 & 0.20 & 0.08 & 0.08 & 66.43 \\
Ours & 0.05 & 0.15 & 0.20 & 0.08 & 0.08 & 66.50 \\
Ours & 0.05 & 0.20 & 0.20 & 0.08 & 0.08 & \textbf{67.24} \\
Ours & 0.10 & 0.20 & 0.20 & 0.08 & 0.08 & 66.23 \\
Ours & 0.20 & 0.20 & 0.20 & 0.08 & 0.08 & 66.12 \\
\hline
Ours & 0.05 & 0.20 & 0.10 & 0.04 & 0.04 & 66.77 \\
Ours & 0.05 & 0.20 & 0.15 & 0.06 & 0.06 & 66.30 \\
Ours & 0.05 & 0.20 & 0.20 & 0.08 & 0.08 & \textbf{67.24} \\
Ours & 0.05 & 0.20 & 0.25 & 0.10 & 0.10 & 67.21 \\
Ours & 0.05 & 0.20 & 0.30 & 0.12 & 0.12 & 66.02 \\

\hline\hline
\end{tabular}
}
\end{table}

\subsection{Temporal Difference Features}
\label{sec:tdf}

Table~\ref{tab:classwise_acc_iou05} shows the accuracy increase per class with temporal difference (TD) features. Large gains are observed for motion sensitive classes such as BasketballDunk, VolleyballSpiking, ThrowDiscus, and CricketBowling. This indicates that temporal difference features help the model to capture the transitional patterns between neighbouring snippets more effectively. Using the TD features, BasketballDunk improves from 53.1\% to 69.6\% and VolleyballSpiking improves from 63.7\% to 78.8\%. We also give additional study on effects of feedback supervision with different feature set in supplementary material Section I(B).
\begin{table}[t]
\centering
\caption{Class wise performance comparison with and without temporal difference (TD) features}
\label{tab:classwise_acc_iou05}

\renewcommand{\arraystretch}{1.08}
\setlength{\tabcolsep}{3.5pt}
\resizebox{0.7\columnwidth}{!}{%
\begin{tabular}{l|ccc}
\hline\hline
Class
& \shortstack{Acc@0.5\\without TD}
& \shortstack{Acc@0.5\\with TD}
& \shortstack{Acc@0.5\\Gain} \\
\hline

BasketballDunk     & 53.1 & 69.6 & +16.5 \\
VolleyballSpiking  & 63.7 & 78.8 & +15.1 \\
ThrowDiscus        & 63.0 & 76.5 & +13.4 \\
CricketBowling     & 65.7 & 77.0 & +11.2 \\
Billiards          & 43.2 & 53.1 & +9.9 \\
CricketShot        & 72.9 & 76.8 & +3.9 \\
TennisSwing        & 76.8 & 79.7 & +2.9 \\

\hline\hline
\end{tabular}
}
\end{table}
\subsection{Limitation and Future Work}
\label{sec:lim}
The dominant localization and long action errors points to specific part of the STAR pipeline that can be further improved. STAR is loaded with All point ranking scheme. This ranking module filters the action candidates on the basis of proposal confidence, neighboring point annotation, and length based constraint. This mechanism seems to be effective for short to medium length proposals, however, for extremely long action instance this mechanism seems to be conservative. Secondly, Rolebank re-ranking also uses the reliable proposal regions to learn the action structure. It is possible that this method fails to learn the structure of extremely long actions and therefore yeilds in the rejection or lower ranking of those extremely long action instances. Therefore, future work will target efficiently designed length prior to long duration classes rather than to apply a uniform penalty mechanism.
\section{Conclusion}
In this paper, we present STAR, a structure-aware framework for point supervised temporal action localization that achieves SOTA results. In STAR, we propose to convert isolated point annotations into richer temporal supervision by learning action start, core, and end roles. The proposed temporal difference interaction with attention module improves snippet level temporal representation. The two proposed ranking schemes i.e. All point ranking and Rolebank re-ranking helps to identify more consistent pseudo proposals with the GT. Furthermore, the feedback from TAD detector to the STAR relay important information about the boundary regions as soft temporal supervision.

The experimental results verify the effectiveness of the proposed STAR method. We verified the results on four standard benchmark dataset. We also verified the results by changing I3D to VideoMAE features and by changing three different TAD detectors i.e. ActionFormer, TriDet, and the recent Mamba. All the experimental results point in a single direction that the idea of estimating action structure information from a single point annotation leads to effective solution for point supervised weakly temporal action localization.
\bibliographystyle{IEEEtran}
\bibliography{STAR_references_completed}

@inproceedings{shou2016temporal,
  author    = {Shou, Zheng and Wang, Dongang and Chang, Shih-Fu},
  title     = {Temporal Action Localization in Untrimmed Videos via Multi-Stage {CNN}s},
  booktitle = {Proceedings of the IEEE Conference on Computer Vision and Pattern Recognition (CVPR)},
  pages     = {1049--1058},
  month     = jun,
  year      = {2016}
}

@inproceedings{zhao2017ssn,
  author    = {Zhao, Yue and Xiong, Yuanjun and Wang, Limin and Wu, Zhirong and Tang, Xiaoou and Lin, Dahua},
  title     = {Temporal Action Detection with Structured Segment Networks},
  booktitle = {Proceedings of the IEEE International Conference on Computer Vision (ICCV)},
  pages     = {2933--2942},
  month     = oct,
  year      = {2017},
  doi       = {10.1109/ICCV.2017.317}
}

@inproceedings{lin2017singleshot,
  author    = {Lin, Tianwei and Zhao, Xu and Shou, Zheng},
  title     = {Single Shot Temporal Action Detection},
  booktitle = {Proceedings of the 25th ACM International Conference on Multimedia (ACM MM)},
  pages     = {988--996},
  month     = oct,
  year      = {2017}
}

@inproceedings{lin2018bsn,
  author    = {Lin, Tianwei and Zhao, Xu and Su, Haisheng and Wang, Chongjing and Yang, Ming},
  title     = {{BSN}: Boundary Sensitive Network for Temporal Action Proposal Generation},
  booktitle = {Computer Vision -- ECCV 2018},
  pages     = {3--21},
  year      = {2018},
  doi       = {10.1007/978-3-030-01225-0_1}
}

@inproceedings{lin2019bmn,
  author    = {Lin, Tianwei and Liu, Xiao and Li, Xin and Ding, Errui and Wen, Shilei},
  title     = {{BMN}: Boundary-Matching Network for Temporal Action Proposal Generation},
  booktitle = {Proceedings of the IEEE/CVF International Conference on Computer Vision (ICCV)},
  pages     = {3889--3898},
  month     = oct,
  year      = {2019}
}

@inproceedings{long2019gtan,
  author    = {Long, Fuchen and Yao, Ting and Qiu, Zhaofan and Tian, Xinmei and Luo, Jiebo and Mei, Tao},
  title     = {Gaussian Temporal Awareness Networks for Action Localization},
  booktitle = {Proceedings of the IEEE/CVF Conference on Computer Vision and Pattern Recognition (CVPR)},
  pages     = {344--353},
  year      = {2019}
}

@inproceedings{xu2020gtad,
  author    = {Xu, Mengmeng and Zhao, Chen and Rojas, David S. and Thabet, Ali and Ghanem, Bernard},
  title     = {{G-TAD}: Sub-Graph Localization for Temporal Action Detection},
  booktitle = {Proceedings of the IEEE/CVF Conference on Computer Vision and Pattern Recognition (CVPR)},
  pages     = {10156--10165},
  year      = {2020}
}

@inproceedings{liu2020pbrnet,
  author    = {Liu, Qinying and Wang, Zilei},
  title     = {Progressive Boundary Refinement Network for Temporal Action Detection},
  booktitle = {Proceedings of the AAAI Conference on Artificial Intelligence (AAAI)},
  volume    = {34},
  number    = {7},
  pages     = {11612--11619},
  year      = {2020}
}

@inproceedings{zhang2024hrpro,
  author    = {Zhang, Huaxin and Wang, Xiang and Xu, Xiaohao and Qing, Zhiwu and Gao, Changxin and Sang, Nong},
  title     = {{HR-Pro}: Point-Supervised Temporal Action Localization via Hierarchical Reliability Propagation},
  booktitle = {Proceedings of the AAAI Conference on Artificial Intelligence (AAAI)},
  volume    = {38},
  number    = {7},
  pages     = {7115--7123},
  year      = {2024},
  doi       = {10.1609/aaai.v38i7.28539}
}

@inproceedings{zhang2022actionformer,
  author    = {Zhang, Chen-Lin and Wu, Jianxin and Li, Yin},
  title     = {{ActionFormer}: Localizing Moments of Actions with Transformers},
  booktitle = {Computer Vision -- ECCV 2022},
  pages     = {492--510},
  year      = {2022}
}

@inproceedings{shi2023tridet,
  author    = {Shi, Dingfeng and Zhong, Yujie and Cao, Qiong and Ma, Lin and Li, Jia and Tao, Dacheng},
  title     = {{TriDet}: Temporal Action Detection with Relative Boundary Modeling},
  booktitle = {Proceedings of the IEEE/CVF Conference on Computer Vision and Pattern Recognition (CVPR)},
  pages     = {18857--18866},
  year      = {2023},
  doi       = {10.1109/CVPR52729.2023.01808}
}

@inproceedings{yang2024dyfadet,
  author    = {Yang, Le and Zheng, Ziwei and Han, Yizeng and Cheng, Hao and Song, Shiji and Huang, Gao and Li, Fan},
  title     = {{DyFADet}: Dynamic Feature Aggregation for Temporal Action Detection},
  booktitle = {Computer Vision -- ECCV 2024},
  pages     = {305--322},
  year      = {2024},
  doi       = {10.1007/978-3-031-72952-2_18}
}

@article{fang2025bdrc,
  author  = {Fang, Zhenying and Yu, Jun and Hong, Richang},
  title   = {Boundary Discretization and Reliable Classification Network for Temporal Action Detection},
  journal = {IEEE Transactions on Multimedia},
  volume  = {27},
  pages   = {5198--5211},
  year    = {2025}
}

@inproceedings{ma2020sfnet,
  author    = {Ma, Fan and Zhu, Linchao and Yang, Yi and Zha, Shengxin and Kundu, Gourab and Feiszli, Matt and Shou, Zheng},
  title     = {{SF-Net}: Single-Frame Supervision for Temporal Action Localization},
  booktitle = {Computer Vision -- ECCV 2020},
  pages     = {420--437},
  year      = {2020},
  doi       = {10.1007/978-3-030-58548-8_25}
}

@inproceedings{lee2021lacp,
  author    = {Lee, Pilhyeon and Byun, Hyeran},
  title     = {Learning Action Completeness from Points for Weakly-Supervised Temporal Action Localization},
  booktitle = {Proceedings of the IEEE/CVF International Conference on Computer Vision (ICCV)},
  pages     = {13648--13657},
  year      = {2021}
}

@inproceedings{yin2023proposal,
  author    = {Yin, Yuan and Huang, Yifei and Furuta, Ryosuke and Sato, Yoichi},
  title     = {Proposal-Based Temporal Action Localization with Point-Level Supervision},
  booktitle = {Proceedings of the 34th British Machine Vision Conference (BMVC)},
  publisher = {BMVA},
  year      = {2023},
  url       = {https://papers.bmvc2023.org/0820.pdf}
}

@inproceedings{xia2024realigning,
  author    = {Xia, Ziying and Cheng, Jian and Liu, Siyu and Hu, Yongxiang and Wang, Shiguang and Zhang, Yijie and Dang, Liwan},
  title     = {Realigning Confidence with Temporal Saliency Information for Point-Level Weakly-Supervised Temporal Action Localization},
  booktitle = {Proceedings of the IEEE/CVF Conference on Computer Vision and Pattern Recognition (CVPR)},
  pages     = {18440--18450},
  month     = jun,
  year      = {2024},
  doi       = {10.1109/CVPR52733.2024.01745}
}

@article{li2024neighbor,
  author  = {Li, Guozhang and Cheng, De and Wang, Nannan and Li, Jie and Gao, Xinbo},
  title   = {Neighbor-Guided Pseudo-Label Generation and Refinement for Single-Frame Supervised Temporal Action Localization},
  journal = {IEEE Transactions on Image Processing},
  volume  = {33},
  pages   = {2419--2430},
  year    = {2024},
  doi     = {10.1109/TIP.2024.3378477}
}

@inproceedings{liu2024smbd,
  author    = {Liu, Mengnan and Wang, Le and Zhou, Sanping and Xia, Kun and Wu, Qi and Zhang, Qilin and Hua, Gang},
  title     = {Stepwise Multi-Grained Boundary Detector for Point-Supervised Temporal Action Localization},
  booktitle = {Computer Vision -- ECCV 2024},
  pages     = {333--349},
  year      = {2024},
  doi       = {10.1007/978-3-031-72667-5_19}
}

@article{zhou2025sdanet,
  author  = {Zhou, Wei and Lin, Kang and Hu, Weipeng and Xie, Chao and Su, Tao and Hu, Haifeng and Tan, Yap-Peng},
  title   = {Snippet-Inter Difference Attention Network for Weakly-Supervised Temporal Action Localization},
  journal = {IEEE Transactions on Multimedia},
  volume  = {27},
  pages   = {3610--3624},
  year    = {2025},
  doi     = {10.1109/TMM.2025.3535336}
}

@inproceedings{liu2025qrot,
  author    = {Liu, Mengnan and Wang, Le and Zhou, Sanping and Xia, Kun and Sun, Xiaolong and Hua, Gang},
  title     = {Boosting Point-Supervised Temporal Action Localization through Integrating Query Reformation and Optimal Transport},
  booktitle = {Proceedings of the IEEE/CVF Conference on Computer Vision and Pattern Recognition (CVPR)},
  pages     = {13865--13875},
  year      = {2025},
  doi       = {10.1109/CVPR52734.2025.01294}
}

@article{fang2026pser,
  author  = {Fang, Zhenying and Hong, Richang},
  title   = {{PseR}: Pseudo-Label Refinement for Point-Supervised Temporal Action Detection},
  journal = {IEEE Transactions on Multimedia},
  volume  = {28},
  pages   = {128--139},
  year    = {2026},
  doi     = {10.1109/TMM.2025.3618538}
}

@inproceedings{yang2026pointboundary,
  author    = {Yang, Junshi and Liu, Shenglan and Sheng, Xuhan and Yan, Gang and Zhou, Yiheng and Feng, Lin and Fan, Jiajun},
  title     = {Bridging the Point to Boundary Gap for Point-Supervised Temporal Action Localization with Single-Stage Inference},
  booktitle = {Pattern Recognition and Computer Vision -- PRCV 2025},
  series    = {Lecture Notes in Computer Science},
  volume    = {16278},
  pages     = {307--320},
  publisher = {Springer},
  year      = {2026},
  doi       = {10.1007/978-981-95-5676-2_21}
}

@article{ma2026temporalconsistency,
  author  = {Ma, Yunchuan and Qing, Laiyun and Li, Guorong and Liu, Yuqing and Qi, Yuankai and Huang, Qingming},
  title   = {Exploring the Temporal Consistency for Point-Level Weakly-Supervised Temporal Action Localization},
  journal = {arXiv preprint arXiv:2602.05718},
  year    = {2026},
  doi     = {10.48550/arXiv.2602.05718}
}

@misc{hrprogithub,
  author = {Zhang, Huaxin and others},
  title = {HR-Pro: Point-Supervised Temporal Action Localization via Hierarchical Reliability Propagation},
  howpublished = {\url{https://github.com/pipixin321/HR-Pro}},
  note = {GitHub repository, accessed: May 14, 2026},
  year = {2024}
}

@misc{videomaev2_tad,
  author       = {{OpenGVLab}},
  title        = {VideoMAE V2: Temporal Action Detection},
  howpublished = {\url{https://github.com/OpenGVLab/VideoMAEv2/blob/master/docs/TAD.md}},
  note         = {Accessed: 2026-08-17}
}

@article{lu2026mambatad,
  author  = {Lu, Hui and Yu, Yi and Lu, Shijian and Rajan, Deepu and Ng, Boon Poh and Kot, Alex C. and Jiang, Xudong},
  title   = {{MambaTAD}: When State-Space Models Meet Long-Range Temporal Action Detection},
  journal = {IEEE Transactions on Multimedia},
  year    = {2026},
  doi     = {10.1109/TMM.2026.3676839}
}

@inproceedings{alwassel2018diagnosing,
  author    = {Alwassel, Humam and Caba Heilbron, Fabian and Escorcia, Victor and Ghanem, Bernard},
  title     = {Diagnosing Error in Temporal Action Detectors},
  booktitle = {Proceedings of the European Conference on Computer Vision (ECCV)},
  pages     = {256--272},
  year      = {2018}
}

@inproceedings{lin2021afsd,
  author    = {Lin, Chuming and Xu, Chengming and Luo, Donghao and Wang, Yabiao and Tai, Ying and Wang, Chengjie and Li, Jilin and Huang, Feiyue and Fu, Yanwei},
  title     = {Learning Salient Boundary Feature for Anchor-Free Temporal Action Localization},
  booktitle = {Proceedings of the IEEE/CVF Conference on Computer Vision and Pattern Recognition (CVPR)},
  pages     = {3320--3329},
  month     = jun,
  year      = {2021}
}

@article{su2021ktmgfn,
  author  = {Su, H. and Zhao, X. and Lin, T. and Liu, S. and Hu, Z.},
  title   = {Transferable Knowledge-Based Multi-Granularity Fusion Network for Weakly Supervised Temporal Action Detection},
  journal = {IEEE Transactions on Multimedia},
  volume  = {23},
  pages   = {1503--1515},
  year    = {2021}
}

@article{li2023superframe,
  author  = {Li, B. and Guo, B. and Zhu, Y. and Yin, J. and Ji, X.},
  title   = {Superframe-Based Temporal Proposals for Weakly Supervised Temporal Action Detection},
  journal = {IEEE Transactions on Multimedia},
  volume  = {25},
  pages   = {3628--3641},
  year    = {2023}
}

@article{chen2023mda,
  author  = {Chen, Z. and Liu, H. and Zhang, L. and Liao, X.},
  title   = {Multi-Dimensional Attention with Similarity Constraint for Weakly-Supervised Temporal Action Localization},
  journal = {IEEE Transactions on Multimedia},
  volume  = {25},
  pages   = {4349--4360},
  year    = {2023}
}

@inproceedings{li2023text,
  author    = {Li, Guozhang and Cheng, De and Ding, Xinpeng and Wang, Nannan and Wang, Xiaoyu and Gao, Xinbo},
  title     = {Boosting Weakly-Supervised Temporal Action Localization with Text Information},
  booktitle = {Proceedings of the IEEE/CVF Conference on Computer Vision and Pattern Recognition (CVPR)},
  pages     = {10648--10657},
  month     = jun,
  year      = {2023}
}

@inproceedings{ren2023pmil,
  author    = {Ren, Huan and Yang, Wenfei and Zhang, Tianzhu and Zhang, Yongdong},
  title     = {Proposal-Based Multiple Instance Learning for Weakly-Supervised Temporal Action Localization},
  booktitle = {Proceedings of the IEEE/CVF Conference on Computer Vision and Pattern Recognition (CVPR)},
  pages     = {2394--2404},
  month     = jun,
  year      = {2023}
}

@inproceedings{rizve2023pivotal,
  author    = {Rizve, Mamshad Nayeem and Mittal, Gaurav and Yu, Ye and Hall, Matthew and Sajeev, Sandra and Shah, Mubarak and Chen, Mei},
  title     = {{PivoTAL}: Prior-Driven Supervision for Weakly-Supervised Temporal Action Localization},
  booktitle = {Proceedings of the IEEE/CVF Conference on Computer Vision and Pattern Recognition (CVPR)},
  pages     = {22992--23002},
  month     = jun,
  year      = {2023}
}

@inproceedings{zhou2023bridging,
  author    = {Zhou, Jingqiu and Huang, Linjiang and Wang, Liang and Liu, Si and Li, Hongsheng},
  title     = {Improving Weakly Supervised Temporal Action Localization by Bridging Train-Test Gap in Pseudo Labels},
  booktitle = {Proceedings of the IEEE/CVF Conference on Computer Vision and Pattern Recognition (CVPR)},
  pages     = {23003--23012},
  month     = jun,
  year      = {2023}
}

@article{moniruzzaman2024f3net,
  author  = {Moniruzzaman, M. and Yin, Z.},
  title   = {Feature Weakening, Contextualization, and Discrimination for Weakly Supervised Temporal Action Localization},
  journal = {IEEE Transactions on Multimedia},
  volume  = {26},
  pages   = {270--283},
  year    = {2024}
}

@inproceedings{yun2024issf,
  author    = {Yun, Wulian and Qi, Mengshi and Wang, Chuanming and Ma, Huadong},
  title     = {Weakly-Supervised Temporal Action Localization by Inferring Salient Snippet-Feature},
  booktitle = {Proceedings of the AAAI Conference on Artificial Intelligence (AAAI)},
  volume    = {38},
  number    = {7},
  pages     = {6908--6916},
  year      = {2024},
  doi       = {10.1609/aaai.v38i7.28516}
}

@inproceedings{li2024dualconfounding,
  author    = {Li, A. and Liu, H. and Sheng, J. and Chen, Z. and Ge, Y.},
  title     = {Efficient Dual-Confounding Eliminating for Weakly-Supervised Temporal Action Localization},
  booktitle = {Proceedings of the 32nd ACM International Conference on Multimedia (ACM MM)},
  pages     = {8179--8188},
  year      = {2024}
}

@inproceedings{lim2024pvlr,
  author    = {Lim, G. and Kim, H. and Kim, J. and Choi, Y.},
  title     = {Probabilistic Vision-Language Representation for Weakly Supervised Temporal Action Localization},
  booktitle = {Proceedings of the 32nd ACM International Conference on Multimedia (ACM MM)},
  pages     = {5507--5516},
  year      = {2024}
}

@inproceedings{ju2021divide,
  author    = {Ju, Chen and Zhao, Peisen and Chen, Siheng and Zhang, Ya and Wang, Yanfeng and Tian, Qi},
  title     = {Divide and Conquer for Single-Frame Temporal Action Localization},
  booktitle = {Proceedings of the IEEE/CVF International Conference on Computer Vision (ICCV)},
  pages     = {13455--13464},
  month     = oct,
  year      = {2021}
}

@article{fu2022crrcnet,
  author  = {Fu, J. and Gao, J. and Xu, C.},
  title   = {Compact Representation and Reliable Classification Learning for Point-Level Weakly-Supervised Action Localization},
  journal = {IEEE Transactions on Image Processing},
  volume  = {31},
  pages   = {7363--7377},
  year    = {2022}
}

@article{fang2023lpr,
  author  = {Fang, Z. and Fan, J. and Yu, J.},
  title   = {{LPR}: Learning Point-Level Temporal Action Localization Through Re-Training},
  journal = {Multimedia Systems},
  volume  = {29},
  number  = {5},
  pages   = {2545--2562},
  year    = {2023}
}

\begin{IEEEbiography}
[{\includegraphics[width=1in,height=1.25in,clip,keepaspectratio]{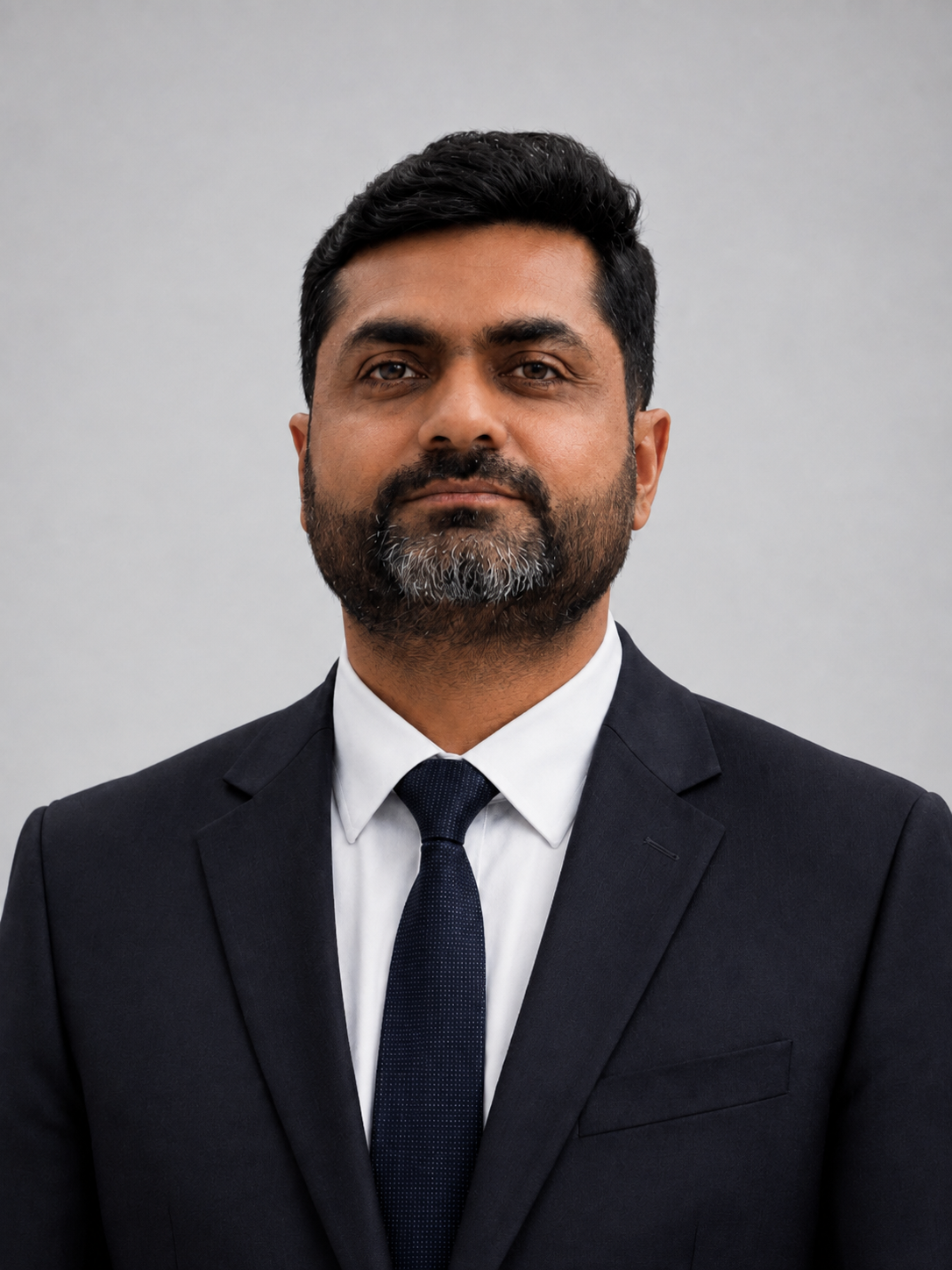}}]
{Yawar Rehman}
received the B.E. degree from Mehran University of Engineering and Technology, Pakistan, and the M.S. and Ph.D. degrees from Hanyang University, South Korea. He served as a Lecturer and subsequently as an Assistant Professor with NED University of Engineering and Technology, Pakistan. He later worked as a Principal Researcher with OnDAI Inc., South Korea. He is currently a Research Professor with Kyungpook National University (KNU), Daegu, South Korea. He has authored/coauthored more than 30 research publications.
\end{IEEEbiography}

\begin{IEEEbiography}
[{\includegraphics[width=1in,height=1.25in,clip,keepaspectratio]{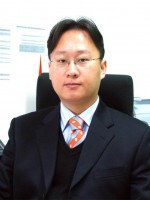}}]
{Min Young Kim}
(Member, IEEE) received the B.S., M.S., and Ph.D. degrees from the Korea Advanced Institute of Science and Technology (KAIST), Daejeon, South Korea, in 1996, 1998, and 2004, respectively. He is currently a Professor with Kyungpook National University (KNU), Daegu, South Korea. He has authored/coauthored more than 180 publications. He is currently also the Director of the KNU LG Electronics Convergence Research Center and the Director of the Research Center for Neurosurgical Robotic Systems, KNU. He also serves as an Associate Editor of the International Journal of Control, Automation, and Systems. His honors include 2024 ICT Express Best Paper Runner Up Award, an Honorable Mention for the Best Paper Award at the 2007 SPIE International Symposium on Optomechatronic Technologies, and the Bronze Award at the 11th Samsung Electronics Human-Tech Thesis Prize in 2004.
\end{IEEEbiography}

\clearpage
\setcounter{section}{0}
\begin{center}

{\LARGE\bfseries Supplementary Material\par}

\vspace{0.3cm}

{\Large\bfseries
STAR: Structure-Aware Temporal Action Recovery
for Point Supervised Temporal Action Localization\par}

\vspace{0.4cm}

{\large
Yawar Rehman, Min Young Kim, Member, IEEE\par}

\vspace{0.3cm}

{\small
This work was submitted to IEEE Transactions on Multimedia
on September 01, 2026.

And uploaded as preprint on personal website
(\url{https://yawar26.github.io/})
while under review status on November 16, 2026.
All rights reserved.\par}

\end{center}

\vspace{0.5cm}

\setcounter{table}{6}
\setcounter{figure}{3}     
\setcounter{equation}{17}

\title{Supplementary Material\\
STAR: Structure-Aware Temporal Action Recovery for Point Supervised Temporal Action Localization}

\author{Yawar~Rehman\,\orcidlink{0000-0001-9743-729X},
	Min~Young~Kim\,\orcidlink{0000-0001-7263-3403}%
    , ~\IEEEmembership{Member,~IEEE}%
}

\IEEEaftertitletext{%
\begin{center}
{\Large\bfseries
This work was submitted to IEEE Transactions on Multimedia on September 01, 2026. And uploaded as preprint on personal website (https://yawar26.github.io/) while under review status on November 16, 2026. All rights reserved.
}
\end{center}
}

\maketitle
\footnotetext{This work has been submitted to the IEEE for possible publication. Copyright may be transferred without notice, after which this version may no longer be accessible.}
\section{Additional Ablation Studies}
\subsection{Ablation Study of STAR with VideoMAE Features}
\label{sec:supp_ablation}
Table~\ref{tab:ablation_study_videomae} reports the ablation experiment using THUMOS'14 videoMAE features. The baseline achieves 62.2\%. The proposed role learning and ranking modules improve the result to 65.8\%. With temporal difference features, the accuracy further increases to 67.1\%. After the feedback supervision, the final accuracy reaches 67.3\%. These results show that the proposed STAR shows almost same trend on both feature sets. And proposed method's components are effective for both I3D features and VideoMAE features.
\begin{table}[!ht]
\centering
\caption{Ablation study of the proposed STAR method with VideoMAE features}
\label{tab:ablation_study_videomae}
\renewcommand{\arraystretch}{1.08}
\setlength{\tabcolsep}{4pt}
\footnotesize
\begin{tabular}{l|cccc}
\hline\hline

Baseline
& \checkmark & \checkmark & \checkmark & \checkmark \\

Role learning and ranking
& & \checkmark & \checkmark & \checkmark \\

Temporal difference features
& & & \checkmark & \checkmark \\

Feedback
& & & & \checkmark \\
\hline

AVG (0.1:0.7)
& 62.2 & 65.8 & 67.1 & 67.3 \\

\hline\hline
\end{tabular}
\end{table}

\subsection{Effect of Feedback Supervision with Different Feature Set}
\label{sec:fb_s}
Table~\ref{tab:feature_feedback_comparison} shows the effect of feedback supervision with I3D and VideoMAE features. With I3D features, feedback supervision improved the detector performance from 66.94\% to 67.24\% and with VideoMAE features the results are improved from 67.05\% to 67.30\%. This consistent gain shows that the proposed method is not limited to only I3D feature representation.
\begin{table}[!ht]
\centering
\caption{Effect of feedback supervision with different feature set}
\label{tab:feature_feedback_comparison}
\renewcommand{\arraystretch}{1.08}
\setlength{\tabcolsep}{6pt}
\begin{tabular}{l|l|c}
\hline\hline
\multicolumn{1}{c|}{Feature} 
& \multicolumn{1}{c|}{Stage} 
& \multicolumn{1}{c}{AVG (0.1:0.7)} \\
\hline

I3D 
& Before feedback 
& 66.94 \\

I3D 
& After feedback 
& 67.24 \\

VideoMAE 
& Before feedback 
& 67.05 \\

VideoMAE 
& After feedback 
& 67.30 \\

\hline\hline
\end{tabular}
\end{table}

\section{Qualitative Analysis}
\label{sec:supp_qualitative}
\begin{figure*}[!ht]
    \centering
    \includegraphics[width=0.76\textwidth]{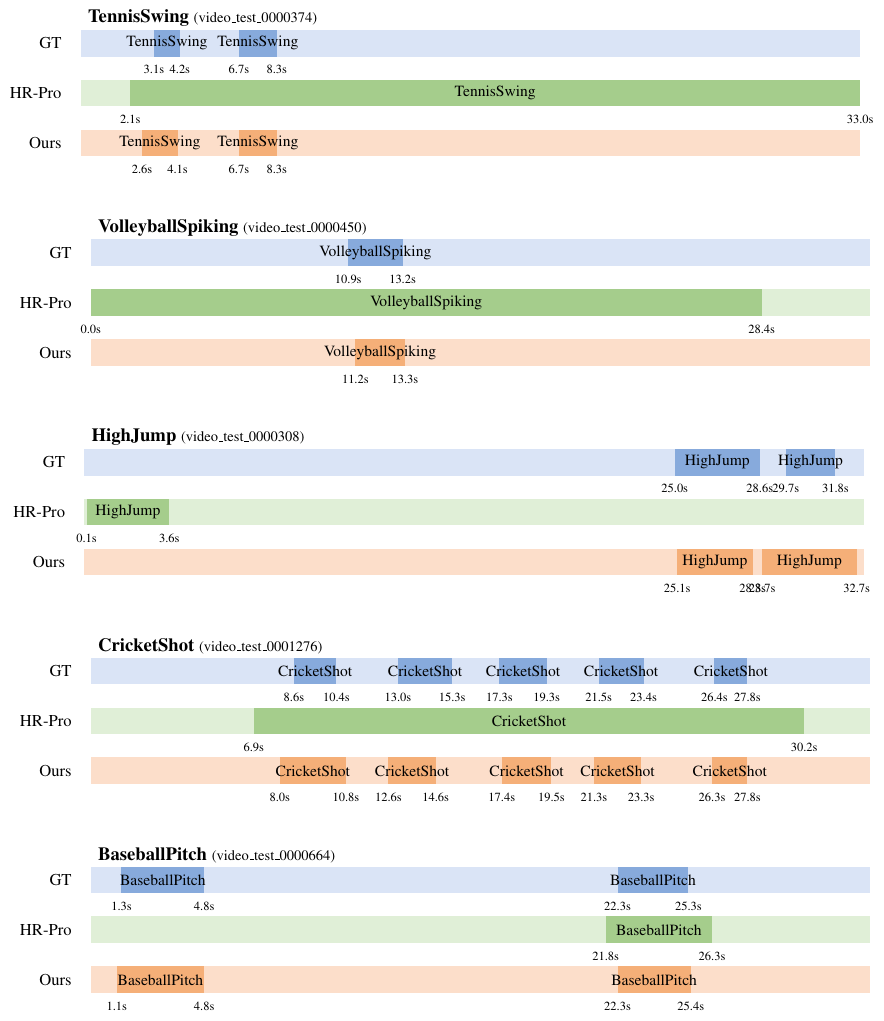}
    \caption{Qualitative comparison of the proposed STAR model on THUMOs'14 dataset. Blue bars show the ground truth for action intervals, green bars shows the performance of fully trained HRPro method, and orange bar show the results of our STAR method with evident improvement. In all classes our STAR method show noticeable improvements. Video names are written instead of pasting images from video for copyright issues}    
    \label{fig:figure3}
\end{figure*}

We present qualitative analysis to show the performance of our proposed method in Fig.~\ref{fig:figure3}. Results are plotted against ground truth (i.e. start and end of action) and with recognized point annotation based action localization method named as HRPro. The HRPro results illustrated here were obtained after training the HRPro model on THUMOS'14 dataset.

In the class TennisSwing, our proposed method STAR localizes both actions accurately. And the second action was localized identical to the ground truth start and end values. In the classes VollyballSpiking, HighJump, and BaseballPitch again our method STAR localized the action with high accuracy. In the CricketShot class there are five shot that has been played by the batsman. HRPro method summarize all the five shots into single action. However, our proposed method STAR cleanly localizes each action with very high localization accuracy. This implies that our method exhibit better temporal separation and boundary classification abilities. 

\begin{figure*}[!ht]
    \centering
    \includegraphics[width=0.8\textwidth]{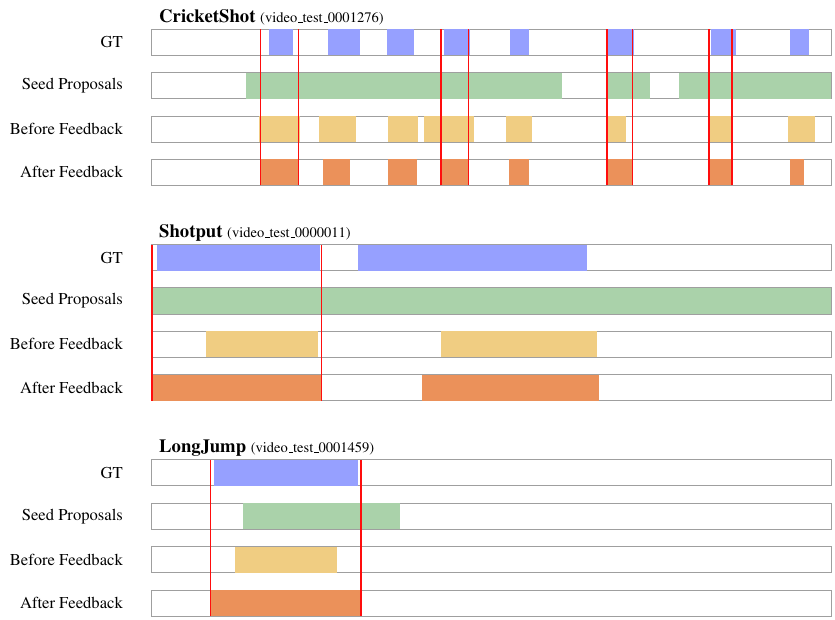}
    \caption{Qualitative analysis of progressive improvements by the proposed STAR method. GT with blue bars depict the ground truth action intervals, Seed Proposals with green bars denote coarse seed proposals generated by the basic S model of HRPro, Before Feedback with yellow bars shows the improvement caused by the STAR method, and After Feedback with orange bars shows the further refined localized action instances after learning the information from feedback. Video names are written instead of pasting images from video for copyright issues}
    \label{fig:figure4}
\end{figure*}

Fig.~\ref{fig:figure4} illustrates the performance of the our proposed method STAR. We present comparison of different stages inside the processing of STAR and we emphasize on how the action proposals are improved to achieve the best action localization using point annotations. In the classes CricketShot, Shotput, and LongJump, the seed proposals are generated by the basic S model using simple classification of action and background. At the before feedback stage STAR improves the proposals by using the proposed temporal difference features, role learning and ranking. At the after feedback stage the proposals are further improved using the relayed information from the temporal action detector back to the STAR network. From the qualitative comparison, it should be noted that the proposed method STAR improves the rudimentary seed proposal with big margin, shows better separated action locations, and temporally better aligns action localizations with the given ground truth start and end locations.

\section{Performance Anaysis with Different TAD Detectors}
\subsection{STAR with Different TAD Detectors}
We further experiment with the proposed method STAR to investigate that its pseudo action labels are better in comparison with state-of-the-art (SOTA) methods for training SOTA temporal action detectors. As shown in Table~\ref{tab:detector_comparison} we evaluate pseudo labels generated by different point annotation based weakly supervised methods and train those with ActionFormer and TriDet TAD detectors. Results show that without any TAD detector LACP obtains 44.5\% and HRPro obtains 51.1\% at AVG@0.3:0.7. When ActionFormer is used, the performance of SOTA method improves, however our STAR method achieves the best results 60.6\% at AVG@0.3:0.7. As compared with PseR w/ HRPro, STAR improves the performance by 3.7 percent. The gain is evident at higher IoU thresholds, where STAR improves mAP@0.5, mAP@0.6, and mAP@0.7 to 64.3\%, 51.6\%, and 34.6\%, respectively. Similar trend can be observed when TriDet is used as TAD detector. STAR improves PseR w/ HRPro by 3.1 percent. The improvements can be seen clearly at higher IoU thresholds.
\begin{table}[t]
\centering
\caption{Performance comparison with different TAD detectors \cite{fang2026pser}. AF denotes ActionFormer.}
\label{tab:detector_comparison}
\renewcommand{\arraystretch}{1.06} 
\setlength{\tabcolsep}{2.6pt} 
\footnotesize

\begin{tabular}{l|c|ccccc|c}
\hline\hline
Method
& TAD
& \multicolumn{5}{c|}{mAP@IoU (\%)}
& Avg. \\
& & .3 & .4 & .5 & .6 & .7 & \\
\hline

LACP
& None
& 64.6 & 56.5 & 45.3 & 34.5 & 21.8
& 44.5 \\

HRPro
& HRPro
& 74.3 & 64.3 & 52.2 & 39.8 & 24.8
& 51.1 \\

\hline

LACP
& AF
& 77.1 & 68.5 & 56.2 & 41.8 & 23.3
& 53.4 \\

HRPro
& AF
& 75.8 & 68.7 & 57.1 & 42.7 & 24.7
& 53.8 \\

PseR-LACP
& AF
& 78.5 & 70.9 & 59.3 & 43.2 & 26.2
& 55.6 \\

PseR-HRPro
& AF
& 78.7 & 70.9 & 60.1 & 45.4 & 29.6
& 56.9 \\

STAR
& AF
& \textbf{79.5} & \textbf{72.9} & \textbf{64.3}
& \textbf{51.6} & \textbf{34.6}
& $\mathbf{60.6}^{\uparrow3.7}$ \\

\hline

LACP
& TriDet
& 78.3 & 69.0 & 56.0 & 42.4 & 24.2
& 54.0 \\

PseR-LACP
& TriDet
& 79.6 & 70.4 & 58.9 & 43.1 & 26.2
& 55.7 \\

HRPro
& TriDet
& 78.5 & 70.9 & 59.6 & 44.1 & 28.3
& 56.3 \\

PseR-HRPro
& TriDet
& \textbf{80.2} & 72.7 & 61.0 & 46.7 & 30.2
& 58.2 \\

STAR (Ours)
& TriDet
& 79.9 & \textbf{72.8} & \textbf{65.5}
& \textbf{52.3} & \textbf{36.1}
& $\mathbf{61.3}^{\uparrow3.1}$ \\

\hline\hline
\end{tabular}
\end{table}

Table~\ref{tab:tad_detector_comparison} shows the comparison of different TAD detector using the STAR pseudo labels. ActionFormer obtains 60.6\%, TriDet improves the result to 61.3\%, and Mamba\cite{lu2026mambatad} achieves the best result at 61.7\%. This indicates that the overall performance for action localization improves with the recent detector. Therefore, we can deduce that our proposed method STAR is TAD detector agnostic and shows reliable temporal supervision and action localizations.
\begin{table}[!ht]
\centering
\caption{STAR performance with different TAD detectors}
\label{tab:tad_detector_comparison}

\renewcommand{\arraystretch}{1.08}
\setlength{\tabcolsep}{6pt}
\scriptsize

\begin{tabular}{l|l|ccccc|c}
\hline\hline
Method
& Detector
& \multicolumn{5}{c|}{mAP@IoU (\%)}
& \shortstack{AVG\\(0.3:0.7)} \\
&
&
0.3 & 0.4 & 0.5 & 0.6 & 0.7
& \\
\hline

Ours
& ActionFormer
& 79.5 & 72.9 & 64.3 & 51.6 & 34.6
& 60.6 \\

Ours
& TriDet
& 79.9 & 72.8 & \textbf{65.5} & \textbf{52.3} & 36.1
& 61.3 \\

Ours
& Mamba
& \textbf{80.3} & \textbf{74.2} & 65.4 & 52.2 & \textbf{36.4}
& \textbf{61.7} \\

\hline\hline
\end{tabular}
\end{table}

\section{Error Analysis}
\label{sec:supp_error}
We perform error analysis using the tool provided by ~\cite{alwassel2018diagnosing} on THUMOS'14 dataset using the I3D features and ActionFormer as temporal action detector (TAD). The analysis is conducted at standard tIoU 0.5 with final STAR predictions that achieves 67.24\%. The analysis reports three diagnostic profiles, i.e. false negative, false positive, and sensitivity profile.

\begin{figure*}[t]
\centering

\begin{minipage}[t]{0.47\textwidth}
\centering
\includegraphics[width=\linewidth]{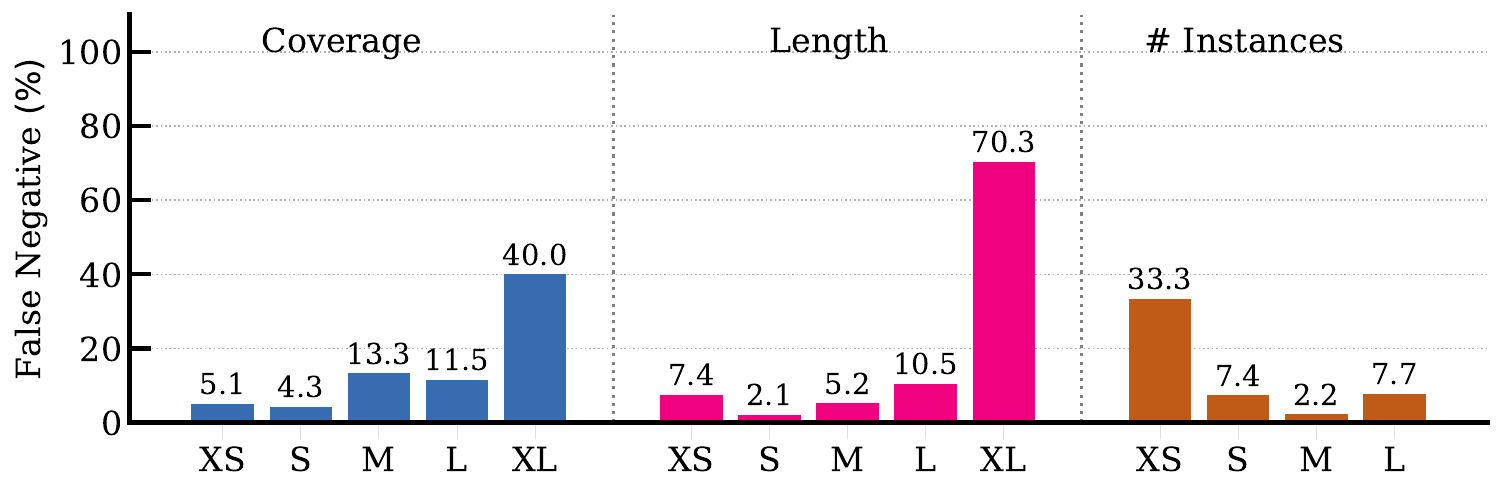}\\[-0.2em]
{\footnotesize (a) FN profile}
\end{minipage}
\hfill
\begin{minipage}[t]{0.47\textwidth}
\centering
\includegraphics[width=\linewidth]{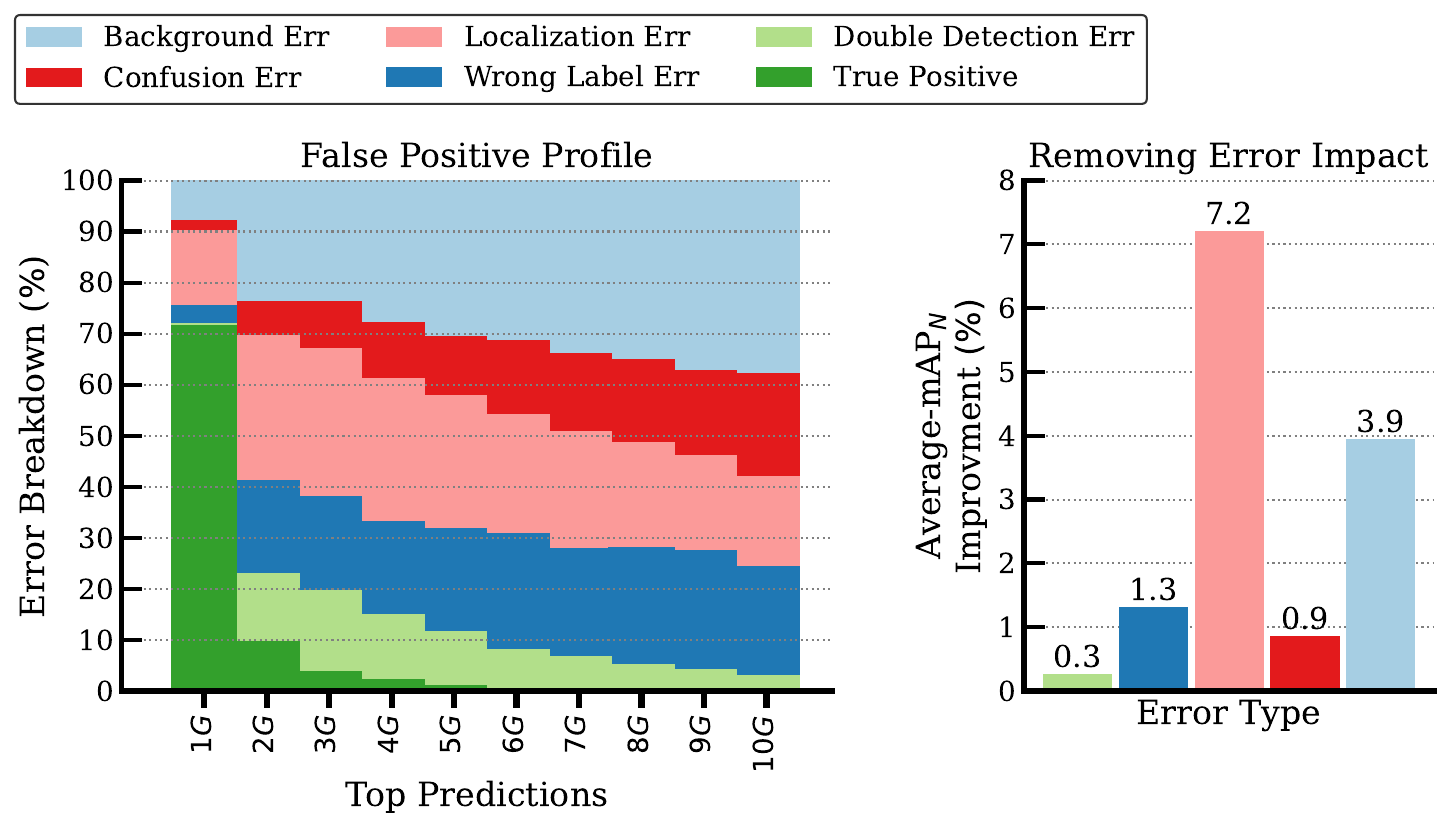}\\[-0.2em]
{\footnotesize (b) FP profile}
\end{minipage}

\vspace{0.8em}

\begin{minipage}[t]{\textwidth}
\centering
\includegraphics[width=\linewidth]{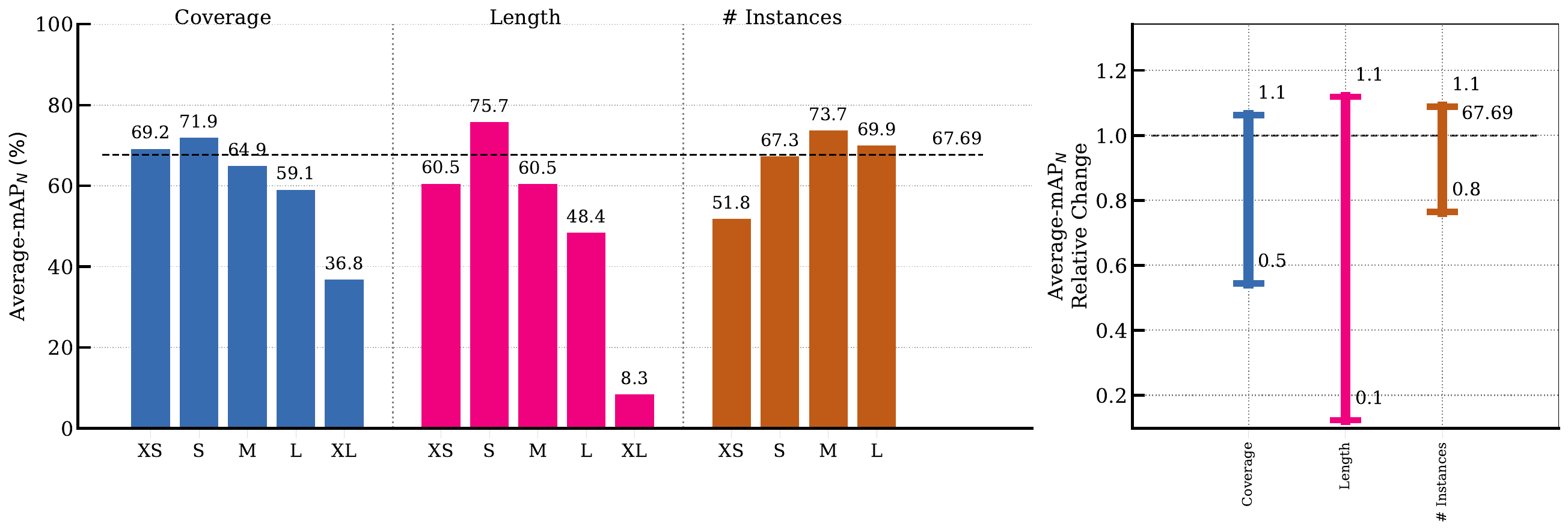}\\[-0.2em]
{\footnotesize (c) Sensitivity profile}
\end{minipage}
\caption{Error analysis of the proposed STAR method using I3D features on THUMOS'14 dataset and ActionFormer as temporal action detector. We report three profiles. (a) The false negative (FN) profile, (b) The false positive (FP) profile, and (c) The sensitivity profile}

\label{fig:sensitivity_i3d}

\end{figure*}

\textit{False Negative Profile:} Fig.~\ref{fig:sensitivity_i3d}(a) shows the false negative (FN) profile. The major miss detection occur in the XL length group that reaches upto 70.3\%. Similar trend is observed for XL coverage videos, where the miss rate is 40.0\%. It implies that our STAR method is highly effective in the range from small to large sized video and action instances, however, in extremely long videos and action instances it struggles and starts to show some weakness. In the number of instances group, miss rate appears in XS. It means that when action instances are limited or less inside a video, the model may produce less confident detections.

\textit{False Positive Profile:} Fig.~\ref{fig:sensitivity_i3d}(b) shows the false positive profile. The most dominant error type is the localization error that is 7.2\%. Removing this localization error will produce the performance improvement in mAP of 7.2\%. Background error is the second largest error that will give the improvement of 3.9\% if it is removed. This implies that the proposed STAR method learns correct action category, however, some predicted segment are misaligned with the GT or include background. 

\textit{Sensitivity Profile:} Fig.~\ref{fig:sensitivity_i3d}(c) shows the sensitivity profile across coverage, action length, and number of instances. Our STAR method performs strongly on XS, S, and M coverage groups, achieving 69.2\%, 71.9\%, and 64.9\% mAP, respectively. However, performance starts to decrease on L and XL coverage groups. A similar trend is observed in the action length, where STAR obtain its best performance on S group at 75.7\% mAP, however the performance decrease sharply on XL group that contains extra long action instances. In contrast, the proposed STAR method is robust to multiple action instance inside a video. This analysis show that STAR is effective for dense, short to medium, and long action localizations. However, extremely long action remains challenging.


\end{document}